%% file: acl_latex.tex
\pdfoutput=1
\documentclass[11pt]{article}

\usepackage[]{acl}

\input{math_commands.tex}

\usepackage{times}
\usepackage{enumitem}
\usepackage{booktabs}
\usepackage{multirow}
\usepackage{latexsym}
\usepackage{graphicx}

\usepackage[T1]{fontenc}
\usepackage[utf8]{inputenc}

\usepackage{microtype}

\usepackage{listings} % For code listing
\usepackage{xcolor}   % For defining colors

\title{DeVAn: Dense Video Annotation for Video-Language Models}

\author{ \bf
Tingkai Liu$^{1}$\thanks{Corresponding: tingkai.liu@columbia.edu}, 
Yunzhe Tao$^{1}$,  
Haogeng Liu$^{2,3}$, 
Qihang Fan$^{2,3}$, 
Ding Zhou$^{1}$,\\
\bf 
Huaibo Huang$^{2}$, 
Ran He$^{2}$, 
Hongxia Yang$^{1}$ \\ \\
$^1$ByteDance, Inc. \\
$^2$MAIS \& CRIPAC, Institute of Automation, Chinese Academy of Sciences, China \\
$^3$School of Artificial Intelligence, University of Chinese Academy of Sciences, Beijing, China \\
}

\begin{document}
\maketitle
\begin{abstract}
We present a novel human annotated dataset for evaluating the ability for visual-language models to generate both short and long descriptions for real-world video clips, termed {\bf DeVAn} (Dense Video Annotation). The dataset contains 8.5K YouTube video clips of 20-60 seconds in duration and covers a wide range of topics and interests. Each video clip is independently annotated by 5 human annotators, producing both captions (1 sentence) and summaries (3-10 sentences). Given any video selected from the dataset and its corresponding ASR information, we evaluate visual-language models on either caption or summary generation that is grounded in both the visual and auditory content of the video. Additionally, models are also evaluated on caption- and summary-based retrieval tasks, where the summary-based retrieval task requires the identification of a target video given \textit{excerpts} of a given summary. Given the novel nature of the paragraph-length video summarization task, we compared different existing evaluation metrics and their alignment with human preferences and found that model-based evaluation metrics provide more semantically-oriented and human-aligned evaluation. Finally, we benchmarked a wide range of current video-language models on DeVAn, and we aim for DeVAn to serve as a useful evaluation set in the age of large language models and complex multi-modal tasks. Code is available at \url{https://github.com/TK-21st/DeVAn}.
\end{abstract}

\section{Introduction}
\input{intro.tex}

\section{Related Work}
\input{related_work.tex}

\section{DeVAn Dataset}
\input{method.tex}

\section{Experiments}
\input{experiment.tex}

\section{Conclusion}
\input{conclusion.tex}

\section{Limitations}
\input{limitations.tex}

\section{Ethics Statement}
\input{ethics.tex}

\subsubsection*{Acknowledgments}
We thank our colleagues Linjie Yang and Heng Wang for their valuable feedback
in the data selection process, and Yiren Jian for porting the VideoCoCa source code to the LAVIS framework.

\bibliography{references}
\bibliographystyle{acl_natbib}

\clearpage
\newpage
\appendix
\input{appendix.tex}

\end{document}

%% file: math_commands.tex
\usepackage{amsmath,amsfonts,bm}

\def\figref#1{figure~\ref{#1}}
\def\Figref#1{Figure~\ref{#1}}

\def\eqref#1{equation~\ref{#1}}
\def\1{\bm{1}}

\DeclareMathAlphabet{\mathsfit}{\encodingdefault}{\sfdefault}{m}{sl}
\SetMathAlphabet{\mathsfit}{bold}{\encodingdefault}{\sfdefault}{bx}{n}

%% file: intro.tex
With billions of active users on video content platforms such as YouTube and TikTok, there has been an unprecedented need for automated complex video understanding. Classically, video understanding has focused on captioning and/or retrieval tasks on short videos with brief (sentence-long) captions. The concise nature of both the videos selected and captions labeled has partly been the result of model limitations, where detailed and nuanced multi-sentence video descriptions have not been possible with lightweight text decoders. With the recent leaps in large language models (LLMs), however, vision-language models (VLMs) now have the opportunity to tap into the immense natural language capabilities of models such as LLaMA \citep{llama,llama2} and ChatGPT\citep{instructgpt,openai2023gpt4}. With tens to hundreds of billions of parameters, these LLMs are able to write entire essays with details and poise that mimic human to an unprecedented extent. With video conversational models such as ImageBind-LLM \citep{han2023imagebindllm}, Video-LLaMA\citep{videollama}, Video-ChatGPT\citep{maaz2023videochatgpt} and VideoChat \citep{li2023videochat} claiming to be able to generate detailed and fine-grained descriptions of video inputs, we believe the time is ripe for an evaluation benchmark that matches the capabilities of modern VLMs powered by LLMs.

\begin{figure*}[!t]
    \centering
    \includegraphics[width=.9\linewidth]{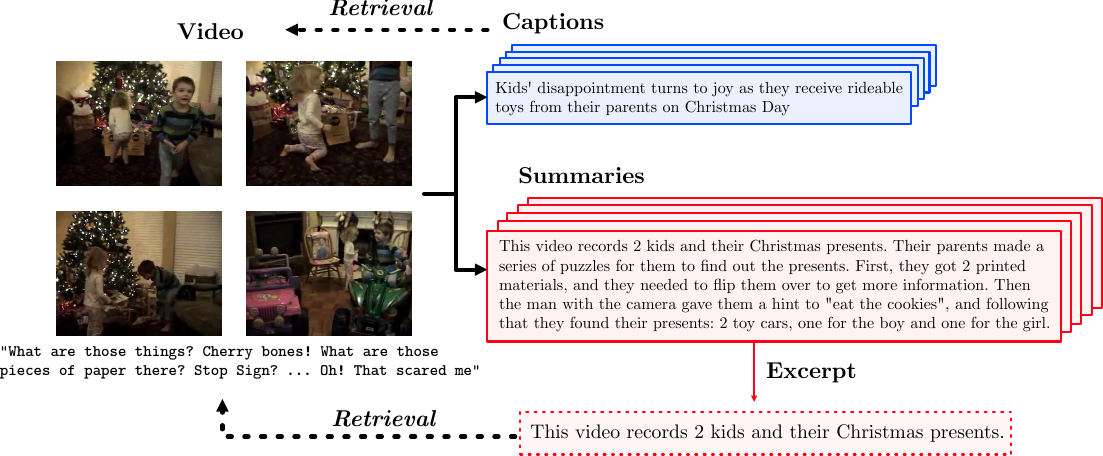} 
    \caption{{\bf Example of DeVAn dataset.} For each video, 5 captions and 5 summaries are independently annotated based on both visual and auditory information of the selected videos. Text-to-video retrievals from video summaries are performed by randomly sampling single-sentence excerpts.
    }
    \label{fig:data_example}
\end{figure*}

In the current work, we focus on videos with multi-shot compositions containing diverse information streams such as dialogues, background music, and complex visual sequences. We developed {\bf DeVAn}, a novel task and dataset for dense long-form video descriptions. This new multi-modal dataset contains 8.5K video clips carefully selected from previously published YouTube-based video datasets (YouTube-8M \citep{abu-el-haija_youtube-8m_2016} and YT-Temporal-1B~\citep{zellers_merlot_2022}) that integrate visual and auditory information. Over the span of 10 months, a team of 24 human annotators (college and graduate level students) created 5 short captions (1 sentence each) and 5 long summaries (3-10  sentences) for each video clip, resulting in a rich and comprehensive human-annotated dataset that serves as a robust ground truth for subsequent model training and evaluation (See \Figref{fig:data_example} for example).

As opposed to short video captions where N-gram based metrics such as CIDEr \citep{vedantam_cider_2015} offer good alignment with human preferences, it is not immediately apparent as how to evaluate long-form video summarization. Additionally, poor instruction-following capabilities of VLMs with frozen LLMs may result in significant variability in the lengths of video descriptions generated, leading to deceivingly low N-gram based metric values even for short video captioning tasks. By comparing N-gram-based metrics and model-based metrics (e.g. BLEURT \cite{sellam_bleurt_2020}, BERTScore \cite{zhang_bertscore_2020}) to human preferences, we find that model-based metrics are better able to capture semantic similarities between model generated responses and human annotations.

Finally, we evaluate different types of VLM architectures on our test set, aiming to provide a comprehensive landscape of what is currently feasible and effective on our task. Specifically, we compare a wide range of recent models featuring frozen LLMs (e.g., VideoChatGPT, ImageBind-LLM, Video-LLaMA) to an end-to-end foundation model developed based on VideoCoCa \citep{videococa}. By training an instance of VideoCoCa using our own training set designed for long-form video summarization, we aim to provide an effective end-to-end baseline model that covers all three tasks at hand. 

Our contributions are summarized as follows:
\begin{itemize}[leftmargin=*]
    \item We introduce a new dataset of human annotated video caption (1 sentence) and summaries (3-10 sentences) to gauge the ability of VLMs to perform long form summary of video content. To the best of our knowledge, DeVAn is the first comprehensive human-annotated evaluation dataset for long-form open-domain video summaries.
    \item We compare different evaluation metrics for long-form video summarization task and find that model-based metrics offer better alignment to human preference.
    \item We evaluate a wide range of video-language models on our DeVAn benchmark including both models with frozen LLMs and end-to-end foundation models, and compare their performances with/without audio information.
\end{itemize}

%% file: related_work.tex
The endeavor to understand and provide textual descriptions of video content has been the subject of numerous research initiatives. We briefly review recent models and datasets relevant to DeVAn.

\paragraph{Video-Language Models} In the video-language model landscape, two model architectures are prevalent. The first category encompasses the \textit{end-to-end trainable models} such as BLIP \citep{li_blip_2022} and VideoCoCa \citep{videococa}, which are designed to learn representations from both videos and text simultaneously, without any frozen modules. In contrast, pioneered by models including BLIP-2 \citep{li_blip-2_2023}, \textit{models with frozen modules} have dominated the video-language scene since the introduction of powerful LLMs like ChatGPT \citep{instructgpt}. Models like BLIP-2 \citep{li_blip_2022}, Video-LLaMA\citep{videollama}, Video-ChatGPT\citep{maaz2023videochatgpt}, and VideoChat \citep{li2023videochat} augment pre-trained frozen linguistic components with additional trainable components, often a lightweight visual backbone. Effectively, such models take advantage of the natural language capabilities of LLMs by providing soft prompts encoded by a lightweight trainable multimodal adaptor. As LLMs are capable of both consuming and generating texts with hundreds if not thousands of words, models in this category are often capable to generate long and detailed video descriptions. We hope that our DeVAn benchmark will contribute the continued advancements in these powerful video-language models.

\paragraph{Video-Language Datasets} Datasets in this domain can be broadly categorized based on their domain specificity and downstream tasks. Refer to Table~\ref{tab:data_stats_avg} for comparison. Under \textit{domain specificity}, datasets like MSVD \citep{msvd}, MSR-VTT \citep{xu_msr-vtt_2016}, YouTube-8M \citep{abu-el-haija_youtube-8m_2016}, YT-Temporal-1B \citep{zellers_merlot_2022}, HD-Vila-100M \citep{hdvila}  provide a panoramic view of diverse video content, fostering a comprehensive model understanding. In contrast, datasets such as How2 \citep{sanabria2018how2} YouCook2 \citep{youcook2} and HowTo100M \citep{howto100m} predominantly focus on instructional content.

In terms of \textit{task orientation}, open-domain datasets mentioned above are often focused on video-to-text generation and retrieval tasks. In contrast, datasets such as Kinetics-700 \citep{kinetics}, ActivityNet \citep{activitynet_original} and ActivityNet Captions \citep{activitynet_captions} focus on more specialized downstream applications such as activity detection.

As our dataset is designed to primarily gauge the ability for models to accurately capture a balanced understanding of overall content and details in a given video, we focused on generation and retrieval tasks for open domain videos.

%% file: method.tex
In this section, we describe the procedure with which DeVAn was constructed and how generation and retrieval task performances are evaluated on DeVAn.

\subsection{Evaluation Dataset} \label{sect:test_set}

The dataset utilized in this study is an amalgamation of YouTube videos, which were source from two previously available large-scale video datasets: YouTube-8M and YT-Temporal-1B. 

The selection of videos for human annotation was focused on English videos with high quality and diversity, and saw one significant evolution during the course of the annotation process. Refer to \Figref{fig:meta_data} in Appendix~\ref{sect:appx:meta} for examples of relevant metadata information used during the video selection process.

\paragraph{First Phase: 2.3K Videos}
In the \textbf{first phase} of the data curation process, videos are selected from YouTube-8M and YT-Temporal-1B datasets based solely video metadata with the following criteria:
\begin{itemize}[leftmargin=*]
    \item Video title, description and subtitles (if applicable) must be primarily in English;
    \item Video must contain Chapter information, which is video keyframe information provided by video uploaders;
    \item Video clips, when segmented based on chapter information, should be between 20 to 60 seconds. 
\end{itemize}
We find that of all videos in the YouTube-8M and YT-Temporal-1B datasets, roughly 1\% satisfied our constraint. Based on the ``category'' metadata information of the videos, we uniformly sampled around 2.3K video segments were curated following this procedure, which form the \textbf{first phase} of our data annotation process.
Note that the 100K training dataset mentioned later in this paper was curated in tandem with the \textbf{first phase} evaluation dataset, as such the distribution of our training dataset aligns best with this portion of the test set (see Section~\ref{sect:train_set} for more information).

\paragraph{Second Phase: 6.2K Videos}
In the \textbf{second phase} of the data annotation process, we adjust the criteria to favor videos for which visual-grounding is \emph{necessary} for accurate annotation. In particular, previous selection criteria (most significantly, the requirement for ``Chapter'' information) led to a bias towards News and Instruction type videos, for which speech information contents were dominant.

\begin{table*}[!t]
\resizebox{\textwidth}{!}{%
\begin{tabular}{@{}lllllllll@{}}
\toprule
Dataset              & Annotation               & Source                   & \begin{tabular}[c]{@{}l@{}}Duration (hrs)\end{tabular} & Domain        & Videos & Clips & Clip Length & Description Length (words) \\ \midrule
HowTo100M             & \multirow{2}{*}{Automatic} & YouTube                        & 134,472                                                  & Instruction & 1.221M & 136M  & 6.5 min         & -                              \\
ActivityNet Captions &                            & YouTube                        & 849                                                      & Open          & 100k   & 100k  & 180 sec         & -                              \\ \midrule
MSVD                 & \multirow{4}{*}{Manual}    & YouTube                        & 4.13                                                     & Open          & -      & 2k    & 10 sec          & 7 (test)                 \\
MSR-VTT              &                            & Commercial Engine & 41.2                                                     & Open          & 7,180  & 10k   & 10-30 sec       & 9.3 (test)               \\
Vatex                &                            & Kinetics-600 + YouTube         & 1,300                                                    & Open          & -      & 41.3k & 20 sec          & 14.5 (test)              \\
YouCook2             &                            & YouTube                        & 175.6                                                    & Instruction & -      & 2k    & 315 sec         & 8.8 (train/val)         \\
\midrule
\textbf{DeVAn}       &           Manual                 & YouTube                        & 95.9                                                     & Open          & 6,709  & 8.5k  & 40.7 sec        & 11 (caption) / 54 (summary) \\ \bottomrule
\end{tabular}
}
\caption{{\bf Video-Language Datasets Comparison.} Refer to \Figref{fig:data_stats} in Appendix~\ref{sect:appx:detailed_stats} for detailed information on DeVAn.}
\label{tab:data_stats_avg}
\end{table*}

As a result, human annotators often heavily favored, for example, the \emph{content} of the News articles being broadcast over the actual \emph{visual setting} of the broadcasting room. This has the undesired consequence that visual-language models with strong language capabilities but weaker visual grounding can potentially have better performance than the more visually-grounded counterparts. Additionally, to avoid videos that are montages of static images, we apply an additional frame-embedding based filter to select videos with high inter-frame variability. The video selection process is as follows.
\begin{enumerate}[leftmargin=*]
    \item The audio content of each video is first processed by \texttt{Whisper-Base} \citep{whisper} to generate automatic speech recognition (ASR) content, followed by an entropy computation, where only videos with entropy lower than 4.2 are kept.
    \item The visual content of each video is evaluated by uniformly sampling 8 frames and computing embedding of each frame using CLIP \citep{clip}; $L_2$ distances between embeddings of neighboring frames are computed and averaged, where only videos with average inter-frame $L_2$ distance above 5.5 are kept.
    \item Video title, description and subtitles (if applicable) must be primarily in English.
    \item Instead of ``Chapter'' information, videos are segmented using key-frames detected via TransNet \citep{transet}. Only segments that satisfy the 20-60 second duration requirement are kept. 
\end{enumerate}
As in the First Phase, the automatically filtered videos were again sampled uniformly based on the ``category'' metadata information. Selected video clips are filtered manually during the annotation process where annotators were provided the option to discard a given video if it is deemed of poor quality: non-English or does not contain sufficient information content for summarization in 3-10 sentences. We find that roughly \emph{20\%} of automatically selected videos were filtered by human annotators.

\begin{figure}[!t]
    \centering
    \includegraphics[width=\linewidth]{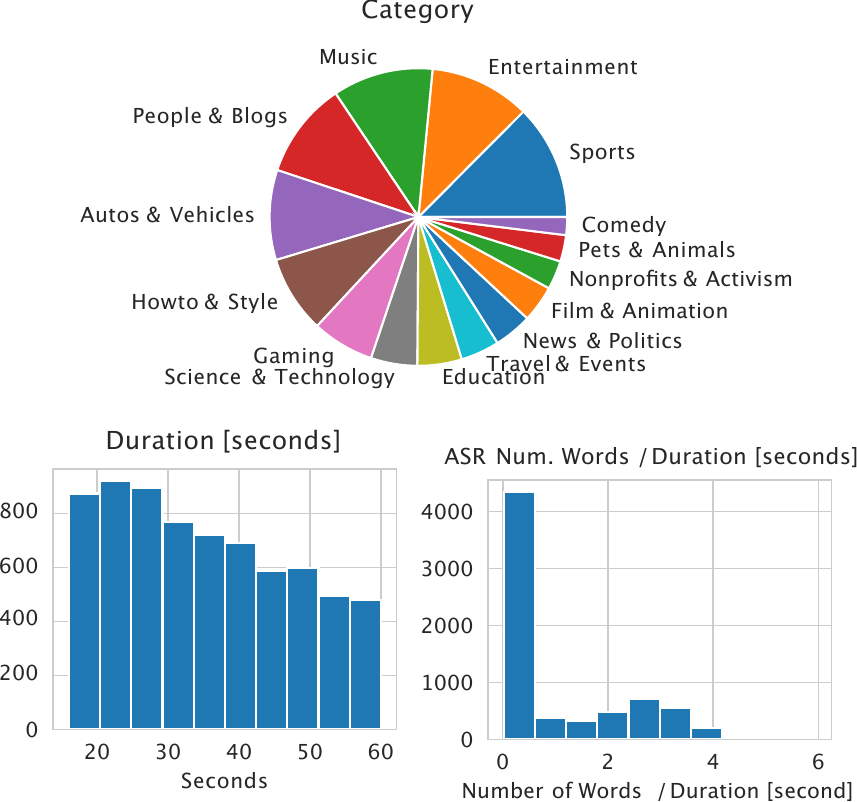} 
    \caption{{\bf Diversity of DeVAn dataset.} Our dataset contains English videos covering a diverse range of topics uploaded across the past 17 years. }
    \label{fig:data_diversity}
\end{figure}

The 10-months-long annotation process is divided into multiple rounds, with each round covering 500-1500 videos. After each round of annotation, 20\% of videos are randomly selected for quality control independent of the original annotators. If systematic problems are detected in a batch of annotations, the entire batch is returned to annotators for revision before going through another round of quality control. In later rounds, as quality of annotation stabilized, the percentage of videos selected for independent quality control is adjusted downwards to a minimum of 7.5\%. This process is repeated until the batch at question is deemed of satisfactory quality, and every batch went through at least one round of revision. Refer to Appendix~\ref{sect:annotation} for more details on the annotation process.

As shown in \Figref{fig:data_diversity}, the final 8.5K evaluation dataset contains videos covering a wide range of topics and interests. The statistics of the videos and annotations are shown in Table~\ref{tab:data_stats_avg} and \Figref{fig:data_stats} in Appendix~\ref{sect:appx:detailed_stats}.
For more qualitative examples of the dataset, refer to Appendix~\ref{sect:examples}.

\subsection{Tasks and Evaluation Metrics}

\paragraph{Video-to-Text Generation Task}
For video-to-text generation tasks, we compared both N-gram-based and model-based evaluation metrics. For N-gram-based metrics, following prior works such as MSR-VTT \citep{xu_msr-vtt_2016}, we report commonly used metrics including BLEU \citep{papineni_bleu_2001}, ROUGE-L \citep{lin_rouge_nodate} and CIDEr \citep{vedantam_cider_2015} to gauge the quality of the model-generated captions. Evaluation of these metrics follow the implementation used in CLIPScore \citep{hessel2021clipscore}, where the Stanford CoreNLP's PTBTokenizer \citep{stanfordcorenlp} is used for text pre-processing. These metrics have proven effective for evaluating the lexical overlap and syntactic structure in brief captions, which are relatively straightforward and independent of language models. In addition to N-gram-based metrics, we also report model-based metric, BLEURT \citep{sellam_bleurt_2020}, which we found to have a better agreement to human preferences (see Section~\ref{sect:metric_alignment} for more details) especially for long-form video summarization tasks. As such, while we report a wide range of evaluation metrics, the model-based BLEURT metric serves as our primary method of evaluation for video-to-text generation tasks.

\paragraph{Text-to-Video Retrieval Task}
For evaluating the efficacy of models in the video retrieval tasks, we follow the classic retrieval accuracy metrics at different levels of granularity: Recall @1, @5, and @10. While this is sufficient for standard text-to-video retrieval tasks using one-sentence video descriptions, it is not directly applicable for retrieval via multi-sentence video summaries. For such task, we introduce a new evaluation methodology where the recall is evaluated using individual sentences from a given video summary. The overall summary-to-video retrieval performance is the averaged recall from each sampled sentences. Note that only sentences with more than 5 words were used to ensure that excerpts of video summaries contain sufficient information for retrieval. This task mimics the common scenario where viewers may desire to search for videos based on memories of partial information. We report Recall @1, @5 and @10 for both caption-to-video and summary-to-video retrieval tasks.

\subsection{Alignment of Evaluation Metrics to Human Preference} \label{sect:metric_alignment}
Given the novel nature of long-form video summarization task, we sought to compare the alignment of different evaluation metrics to human preferences.

To start, we computed the Spearman Rank Correlation between different evaluation metrics on both captioning and summarization tasks as shown in \Figref{fig:human_metric_corr}. Note that Spearman Rank Correlation was chosen over Pearson Correlation to emphasize pairwise consistency between evaluation metrics. Using one annotated response as prediction and all other responses as ground truths, we computed N-gram based (BLEU-4, ROUGE-L, CIDEr) and model-based (BLEURT) metrics for both captioning and summarization tasks. We observe that across all annotators and all videos, the correlation between N-gram-based and model-based metrics is 0.65 for captioning task, while the minimum correlation within N-gram metrics is 0.73. This result suggests that N-gram based metrics may offer more consistent evaluation for captioning task. In contrast, N-gram metric such as CIDEr appear poorly correlated with other metrics for long-form video summarization tasks.

To determine the most suitable metric for evaluating long-form summarization task, we intuited that the alignment of a metric to human preferences can be measured by the metric's ability to tightly cluster annotations of the same video created by different labelers. Formulating this intuition as a text-to-text retrieval problem, we compared recall performances of identifying the same annotations from different annotators using different evaluation metrics, and found that while CIDEr and BLEURT have similar recall performances for video captions, BLEURT significantly outperforms all N-gram based metrics for video summaries by over 14\% (see Table~\ref{tab:text_to_text_ret} in Appendix~\ref{sect:appx:human_perf_details}). This provides indirect support that the model-based BLERUT metric may be better aligned to human preferences especially for long-form video summaries. To further validate this result, we randomly selected 20 videos and manually ranked summaries from annotators 1 and 5 by their perceived quality and relatedness to the video content. We than compared this human labeled ranking result to the ranking created by aforementioned metrics, computed using label 1/5 as predictions and label 2/3/4 as ground truths.
We observe a 56\% alignment of human ranking to ranking by CIDEr score and a 67\% alignment to that by BLEURT, reinforcing the previous finding that BLEURT is better aligned to human preferences.

It is worth noting that while the difference in human alignment between BLEURT and CIDEr appears significant, it is biased by preferences by human labelers and limited by the number of videos sampled during comparison. Nevertheless, the difference between CIDEr and BLEURT for video summaries point to the differences between semantic and lexical similarities when evaluating paragraph-length textual descriptions. We chose BLEURT as our metric for video summarization task but it remains unclear how better alignment to human preferences should be evaluated and implemented for complex multi-modal tasks with long form text generation.

\begin{figure}[!t]
    \centering
    \includegraphics[width=.9\linewidth]{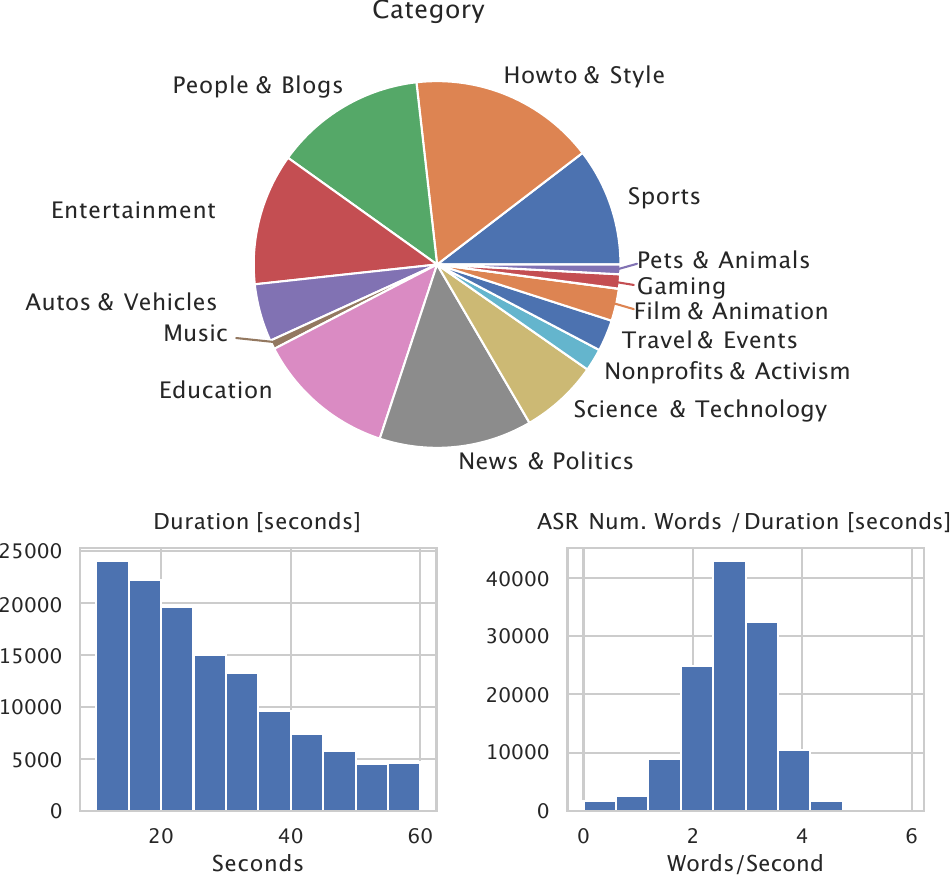} 
    \caption{{\bf Diversity of training dataset.} Our training dataset contains captions and summaries for 100K ASR-rich video segments. Note that as opposed to test dataset in \Figref{fig:data_diversity}, the ASR Number of Words normalized by video duration does not have 
    a significant concentration around 0, indicating that all videos in the training dataset contains a significant amount of ASR information.
    }
    \label{fig:training_data_stats}
\end{figure}

\begin{table*}[!t]
\centering
\resizebox{\textwidth}{!}{%
\begin{tabular}{@{}cccccccccccccccc@{}}
\multirow{3}{*}{Model}            & \multirow{3}{*}{Audio} & \multicolumn{7}{c}{\textbf{Caption}}                                                                         & \multicolumn{7}{c}{\textbf{Summary}}                                                                         \\
                                  &                        & \multicolumn{4}{c}{\textit{Generation Metrics}}             & \multicolumn{3}{c}{\textit{Retrieval Metrics}} & \multicolumn{4}{c}{\textit{Generation Metrics}}             & \multicolumn{3}{c}{\textit{Retrieval Metrics}} \\
                                  &                        & B4           & R             & C            & B-RT          & R@1            & R@5           & R@10          & B4           & R             & C            & B-RT          & R@1            & R@5           & R@10          \\ \midrule
Human (Avg)                       & Raw                    & 6.3          & 32.1          & 53.9         & 50.5          & -              & -             & -             & 15.7         & 34.5          & 36.9         & 55.6          & -              & -             & -             \\
Human (Min)                       & Raw                    & 4.5          & 29.5          & 47.1         & 48.6          & -              & -             & -             & 12.4         & 32.1          & 30.9         & 53.6          & -              & -             & -             \\ \midrule
ImageBind-LLM                     & N/A                    & 0.3          & 20.0          & 2.1          & 34.0          & -              & -             & -             & 1.5          & 22.7          & 1.1          & 45.8          & -              & -             & -             \\
\multirow{2}{*}{Video-LLaMA2 13B} & N/A                    & 0.1          & 7.9           & 0.0          & \textbf{47.2} & -              & -             & -             & 0.5          & 18.2          & 0.0          & 39.9          & -              & -             & -             \\
                                  & Raw                    & 0.1          & 7.9           & 0.0          & 47.1          & -              & -             & -             & 0.5          & 18.2          & 0.0          & 40.0          & -              & -             & -             \\
\multirow{2}{*}{Video-LLaMA2 7B}  & N/A                    & 0.1          & 10.8          & 0.0          & 43.6          & -              & -             & -             & 0.5          & 19.1          & 0.0          & 43.9          & -              & -             & -             \\
                                  & Raw                    & 0.1          & 10.8          & 0.0          & 43.6          & -              & -             & -             & 0.5          & 19.1          & 0.1          & 43.9          & -              & -             & -             \\
VideoChatGPT                      & N/A                    & 0.4          & 19.9          & 2.0          & 40.5          & -              & -             & -             & \textbf{2.9} & \textbf{24.4} & \textbf{5.8} & \textbf{46.7} & -              & -             & -             \\ \midrule
\multirow{2}{*}{VideoCoCa}        & N/A                    & 0.2          & 13.2          & 2.3          & 17.6          & 32\%           & 50\%          & 58\%          & 0.9          & 16.4          & 3.3          & 23.9          & 25\%           & 41\%          & 48\%          \\
                                  & ASR                    & \textbf{0.8} & \textbf{20.3} & \textbf{9.2} & 21.9          & \textbf{36\%}  & \textbf{53\%} & \textbf{59\%} & 2.0          & 21.6          & 5.5          & 22.9          & \textbf{27\%}  & \textbf{42\%} & \textbf{48\%} \\ \bottomrule
\end{tabular}
}
\caption{{\bf Results for generation and retrieval tasks of DeVAn evaluation dataset.} For evaluation of human performance, annotation from each annotator is used as \textit{prediction} and computed against \textit{ground truth} results from all other 4 annotators. The overall metrics are then aggregated via Average and Minimum. 
Note that only results for VideoCoCa models are shown for retrieval tasks, since other VLMs tested in the current work do not supported retrieval.
{\bf B4}: BLEU-4, {\bf R}: Rouge-L, {\bf C}: CIDEr, {\bf B-RT}: BLEURT. {\bf R@k}: Recall at k.}
\label{tab:results_gen_ret}
\end{table*}

\subsection{Training Dataset} \label{sect:train_set}
In addition to the 8.5K evaluation dataset, we also prepared a training dataset with 100K video clips whose captions and summaries were generated using metadata information (e.g., title, description, category, ASR, etc.). The training dataset is comprised of 100,000 videos segments, selected via the same procedure as in the {\bf first phase} of evaluation dataset curation (see Section~\ref{sect:test_set}). Note due to time constraint, we were not able to collect another training set based on the selection criteria in the {\bf second phase} of evaluation dataset development. Consequently, the training dataset is skewed towards ASR-rich videos (see \Figref{fig:training_data_stats}).

Similar to videos in the evaluation dataset, videos in the training dataset were segmented based on the video chapter information. A prompt template was then used to create queries for an LLM to generate 5 captions and summaries for each video (see Appendix~\ref{appx:sect:prompt}). To exercise control over the text generation process and maintain consistency, each prompt was prefixed with ``This video''. At the time of dataset creation, the \texttt{gpt-3.5-turbo} \citep{instructgpt} model was observed to have significantly more issues related to hallucinations as compared to \texttt{text-davinci-003} \citep{gpt3}, thereby motivating the choice of the seemingly less powerful but more ``reliable'' \texttt{text-davinci-003} for our specific requirements.

It is important to note that the main focus of the current work is on the \textit{zero-shot} performances of current models on our DeVAn \textit{evaluation} dataset, and, as later discussed in Section 4.3, only the end-to-end VideoCoCa model was fine-tuned on the DeVAn training dataset. This is due to the fact that our VideoCoCa model was initialized from the CoCa-ViT-L-14 ckpt (see Section~\ref{sect:end_to_end} and Appendix~\ref{sect:appx:training_details} for more details), which was originally trained on COCO-styled short captions. Direct inference of such model for video summarization tasks resulted in catastrophically poor performance as it lacks the capability to be prompted via natural language to generate video descriptions of different lengths (see Appendix~\ref{sect:supp:videococa_no_training}). However, as the DeVAn training dataset was created using only the metadata information of the videos, the training dataset differ significantly from the evaluation dataset in terms of category, audio content, description length, and vocabulary, thereby constituting a zero-shot evaluation of the VideoCoCa model.

%% file: experiment.tex
In this section, we describe both human performance and model performance on the DeVAn evaluation set.
To comprehensively evaluate the capabilities and limitations of various architectures for video captioning and retrieval, we consider two distinct types of models, each with its own set of advantages and drawbacks.

\begin{table}[!t]
    \centering
    \small
    \begin{tabular}{cccc}
    \toprule
    \multirow{2}{*}{Task} & \multirow{2}{*}{Comparison} & \multicolumn{2}{c}{Metric} \\
    \cmidrule(lr){3-4}
    & & C & B-RT \\
    \midrule
    \multirow{2}{*}{Caption} & Across Annotators & 31 & - \\
    & Across Time & 41 & - \\
    \midrule
    \multirow{2}{*}{Summary} & Across Annotators & 50 & 60 \\
    & Across Time & 38 & 56 \\
    \bottomrule
    \end{tabular}
    \caption{{\bf Same annotator re-labeling summaries of the same video twice is equivalent to 
    the same video summary labeled by different annotators.} 150 randomly selected captions and summaries were relabeled by the same annotator at least 1 month apart from the original annotation. We observe that annotations by the same annotator 1 month apart have similar level of discrepancy as compared  to annotations by different annotators.
    }
    \label{table:cap_sum_redo}
\end{table}

\subsection{Human Performance}
To establish a human performance benchmark, we use a strategy similar to that described in Section~\ref{sect:metric_alignment} where one human annotation is used as the ``prediction'', while the remaining four annotations are used as ``references''. The aggregated results across annotators for both captioning and summarization tasks are shown in Table~\ref{tab:results_gen_ret}. For detailed metrics of each annotator, see Table~\ref{table:human_cap_sum} in Appendix~\ref{sect:appx:human_perf_details}.

While the above evaluation establishes the consistency of human annotations across annotators, we also wanted to evaluate the consistency across time. To that end, we compared similarities of annotations for the same video created by the same annotator more than one month apart. A shown in Table~\ref{table:cap_sum_redo}, compared to other annotators, same annotator is able to achieve a higher CIDEr score for captioning task but a lower CIDEr and BLEURT score for summarization task. This indicates that annotators are able to maintain a much higher consistency for captioning task over summarization task, which is reasonable given the extensive nature of long-form video summarization. In fact, comparison of BLEURT scores of summarization in Table~\ref{table:cap_sum_redo} indicates that annotation consistency \emph{across time} is similar that \emph{across annotators}, suggesting the high subjective difficulty of the long-form video summarization task. It is therefore crucial to have multiple (5 in our case) summary annotations for each video to ensure the diversity of ground truths.

\subsection{Trainable Visual Encoder with Adaptor on Frozen LLMs}
Recently, many methods combining visual encoders with frozen LLMs have emerged. Pioneered by models including BLIP-2, recent additions to this category of models include ImageBind-LLM, Video-LLaMA2 and VideoChatGPT, which all involve projecting video encoding (visual and optionally auditory information) into soft tokens that serve as prefix for frozen LLMs. We performed zero-shot instruction-based evaluations of these models with default system prompts from demos in their corresponding repositories, and using instructions \textit{``Describe this video in ONE sentence.''} for captioning task and \textit{``Summarize this video in three to ten sentences.''} for summarization task.

\begin{figure}[!t]
    \centering
    \includegraphics[width=\linewidth]{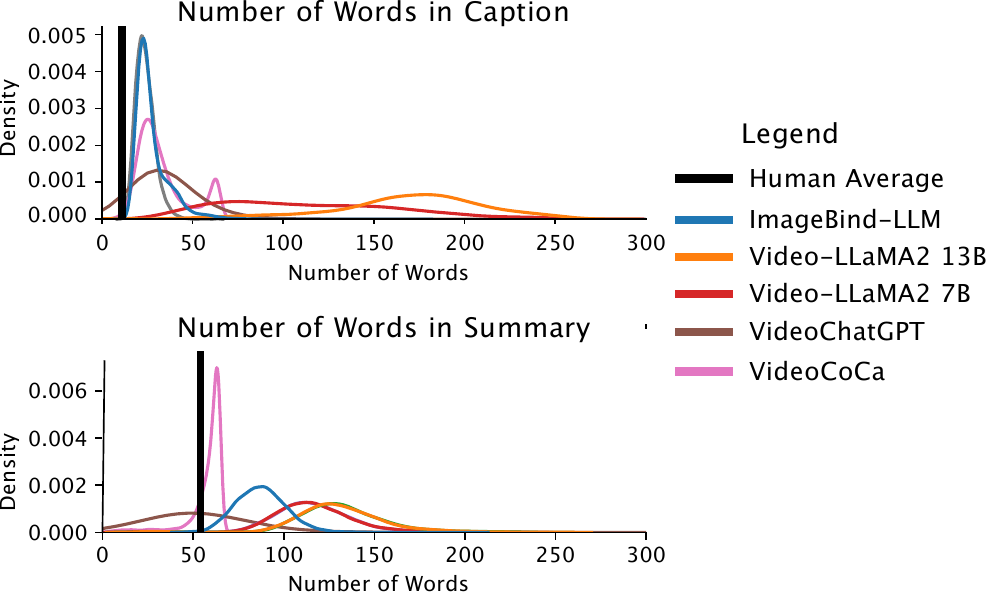} 
    \caption{{\bf Distribution of Number of Words in Captions and Summaries.} Note that for legibility, only distributions for models without audio signals were shown. However, we found that the distributions of caption and summary lengths do not vary significantly with the introduction of audio signals.
    }
    \label{fig:dist_cap_sum_words}
\end{figure}

Surprisingly, as shown in Table~\ref{tab:results_gen_ret}, we found that models like Video-LLaMA2 have very poor video-to-text generation performances for both captioning and summarization when evaluated using N-gram-based metrics. Upon closer examination of the generated results, we realize that due to poor instruction-following (or a lack of prompt engineering), models with frozen LLMs generated responses with highly variable length as shown in \Figref{fig:dist_cap_sum_words}. Pathological examples of such cases with close to 0 CIDEr but high BLEURT scores are shown in Appendix~\ref{sect:pathological_captions}, which all corresponding to very long, sometimes repeating, generated captions. As such, while the results generated are semantically related to the video and human annotations (as found both via human evaluation and BLEURT), they have very low performances in N-gram-based evaluations.

\subsection{End-to-End Foundation Model}\label{sect:end_to_end}
\begin{figure}[!t]
    \centering
    \includegraphics[width=\linewidth]{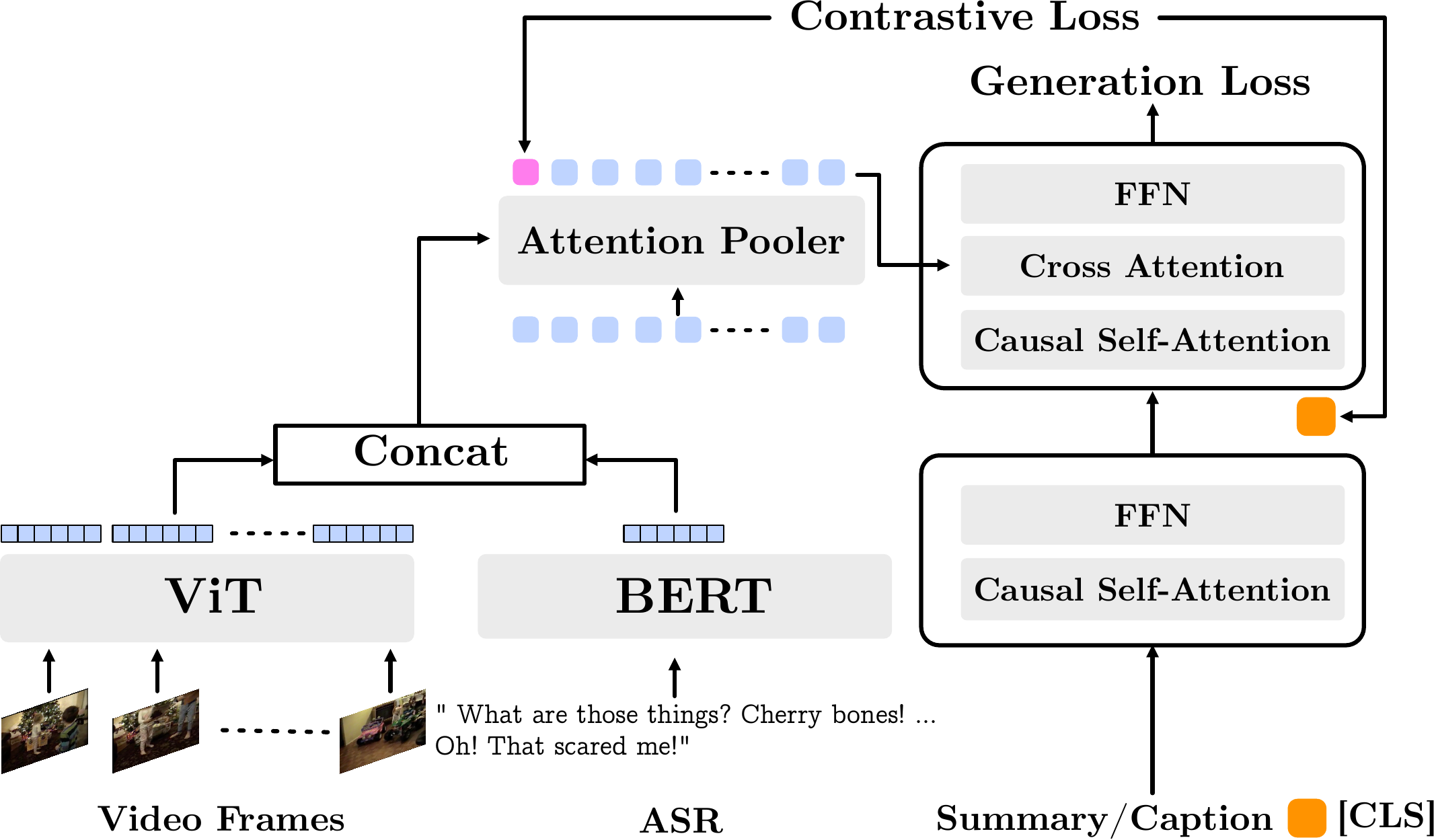} 
    \caption{{\bf End-to-End Model Architecture.} Our End-to-End model combines VideoCoCa
    model architecture with an additional ASR encoder. Frame-level embeddings of VideoCoCa
    and ASR embeddings are concatenated before passing through the Attention Pooler.
    Note that the contrastive loss is computed using the first output embedding of the 
    Attention Pooler.}
    \label{fig:model}
\end{figure}

Despite recent success of VLMs with frozen LLMs, they often cannot support retrieval tasks due to the lack of a dedicated language encoder. To provide a baseline model that is able to achieve competitive results in both video-to-text generation and text-to-video retrieval tasks, we developed an end-to-end trainable foundation model based on the VideoCoCa architecture \citep{videococa} that includes both a visual encoder and ASR encoder (see \Figref{fig:model} for model architecture). As no open source VideoCoCa implementation was available, we created an implementation from scratch following the {\bf Attention Pooler} type model described in the VideoCoca manuscript. 

Following CoCa \citep{yu_coca_2022} and VideoCoCa, the text encoder takes in Caption or Summary as input where a special \texttt{[CLS]} token is suffixed to all input sequences. The text encoder is evenly divided into unimodal (bottom) and multi-modal parts, where the output encoding of the \texttt{[CLS]} token by the unimodal encoder is used to compute contrastive training objective against other modalities.

On the visual encoder side, 8 frames are uniformly sampled from each video and encoded independently by ViT. The output visual encoding of each frame is concatenated to form the overall representation of the visual information in an input video.
The visual encoder output is further concatenated with BERT encoding of the ASR information. The encodings are then integrated and compressed by the ``Attention Pooler'' module with 256 query tokens. The first output token of Attention Pooler is used to compute contrastive loss against encoding of \texttt{[CLS]} token mentioned above.

Following CoCa, the training objective is a combination of generation loss and contrastive loss. All parameters are initialized from the OpenCLIP \citep{openclip} implementation of CoCa, except for the ASR encoder which was initialized from \texttt{BERT-base} \citep{devlin2019bert}. Results of the video-to-text generation and text-to-video retrieval on DeVAn can be found in Table~\ref{tab:results_gen_ret}. We note that the integration of audio information via ASR significantly improves both retrieval and generation performances of our model. This is consistent with the observation that roughly 25\% of the evaluation dataset contains videos rich in speech content.

%% file: conclusion.tex
In this paper, we introduce a new multi-modal dataset curated from a diverse range of YouTube videos, designed to gauge the capabilities of visual-language models in long-form video summarization tasks. Through a carefully orchestrated annotation process involving multiple human annotators, multiple rounds of video selection and quality control, we ensure that the dataset is comprised of high-quality and diverse captions and summaries. 

We show that while qualities of one-sentence captions can be accurately evaluated by N-gram, multi-sentence summaries require a more semantically aligned metric such as BLEURT. Using such metrics, we provide an extensive benchmarking of the current state-of-the-art video-to-text generation models. However, current visual-language models with frozen LLMs often do not support retrieval tasks due to lack of language encoders. To provide a baseline for text-to-video retrieval tasks, we finetuned VideoCoCa using our automatically generated training dataset, which achieved reasonable retrieval, but comparatively weaker generation performances.

As the field of video summarization and captioning continues to evolve, it is imperative that datasets and evaluation metrics keep pace. Our work aims to serve as a stepping stone in this direction, providing a balanced approach to video understanding.

%% file: limitations.tex
While DeVAn strives to become a comprehensive and objective evaluation benchmark for video-to-text generation and text-to-video retrieval tasks, certain limitations warrant consideration.
\begin{itemize}[leftmargin=*]
    \item A limitation of the current work both in the creation of training dataset and in the bench-marking of existing models is that of prompt engineering. It has long been observed that LLMs can be highly sensitive to prompt design \cite{prompt_engineering}, which may contribute to the high variability in the length of the generated responses. While we tried to mitigate this problem by staying close to the default instructions in used in the corresponding works, we still observed problems of poor instruction following.
    \item Additionally, despite an effort to compare different available metrics, neither CIDEr nor BLEURT achieved higher than 80\% alignment to human preferences. While it is feasible to pose the problem in a multiple choice format to mitigate this problem, this would usually require finetuning models to following MCQ instructions. Alternatively, a Reward Model for video annotations may be trained based on human preferences (similar to finetuning BLEURT). However, this would require significantly more human annotations which was unfortunately infeasible.
\end{itemize}
Due to time and resource constraints, these limitations were not addressed in the current work. However, we do expect performance gains of the evaluated models by addressing these limitations.

%% file: ethics.tex
This research aims to provide an objective and comprehensive benchmark for video-to-text generation and text-to-video retrieval tasks from open domain videos.
Although our research does not involve human subjects directly, it is important to acknowledge and discuss the broader ethical implications.

\paragraph{Data Bias and Fairness:} Both training and testing videos in DeVAn are selected from YouTube-8M and Youtube-Temporal-1B datasets, and follow their data curation and anonymization practices. While these datasets are widely used, we acknowledge that we cannot fully ascertain the extent to which they may contain discriminatory, biased, or sensitive material. Given that our finetuned VideoCoCa model inherits the biases present in our training datasets, there exists the risk of perpetuating or even amplifying existing societal biases. Despite the broad acceptance of these datasets, caution should be exercised.

\paragraph{Responsible Usage:} Like all open domain video datasets, it is imperative to implement safety mechanisms to ensure that models trained or evaluated on our dataset do not inadvertently produce outputs that could disclose sensitive or personal information.

\paragraph{AI Assistant Usage:} ChatGPT was used for grammatical correction in the current manuscript.

%% file: appendix.tex
\section{YouTube Metadata information}\label{sect:appx:meta}
\begin{figure}[h]
    \centering
    \includegraphics[width=\linewidth]{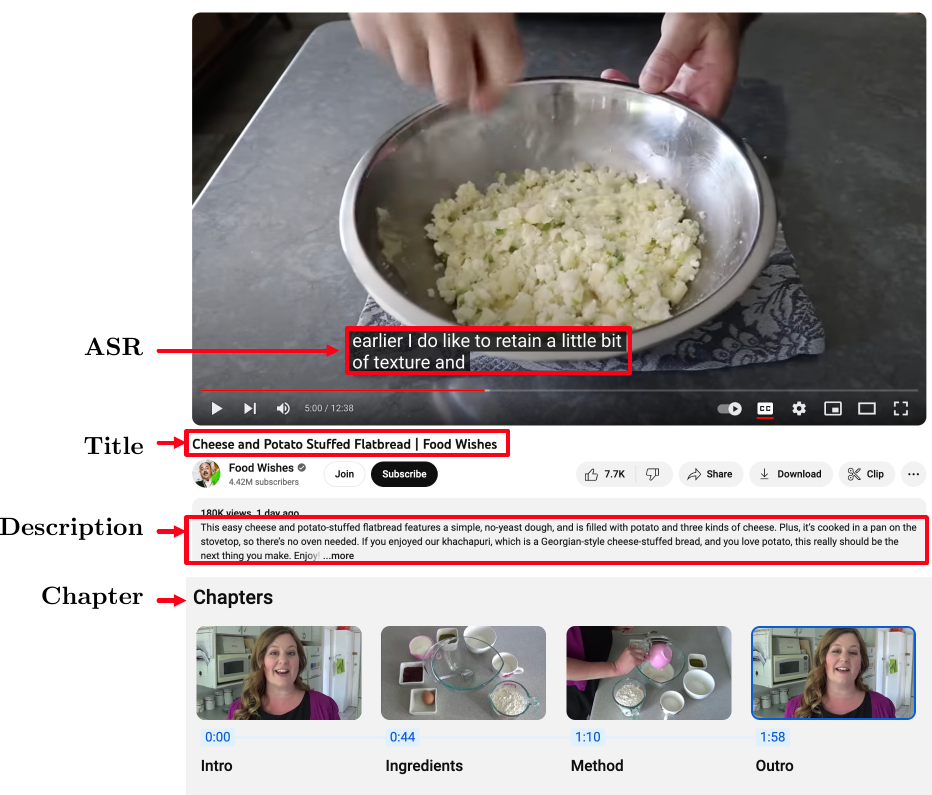}
    \caption{{\bf Example of YouTube metadata information.} 
        Note that only metadata that are relevant for video selection is shown. ``Chapter'' information contains 
        keyframe information with headers provided by the video uploader.
    }
    \label{fig:meta_data}
\end{figure}

YouTube metadata information refers to all information related to an uploaded video except for the video and audio content. This includes information such as title, description, category, tags, asr, chapter, playlist, subtitles, etc. In \figref{fig:meta_data}, we highlight a few important metadata information that is relevant for video selection in our work. 

We highlight ``Chapter'' information, which corresponds to keyframe information with human annotated segment subtitles. These chapter information naturally divide video into semantically meaningful chunks that we used to split longer videos into 20-60 second clips.

\section{Detailed Statistics of DeVAn Evaluation Dataset}\label{sect:appx:detailed_stats}
\begin{figure}[h]
    \centering
    \includegraphics[width=\linewidth]{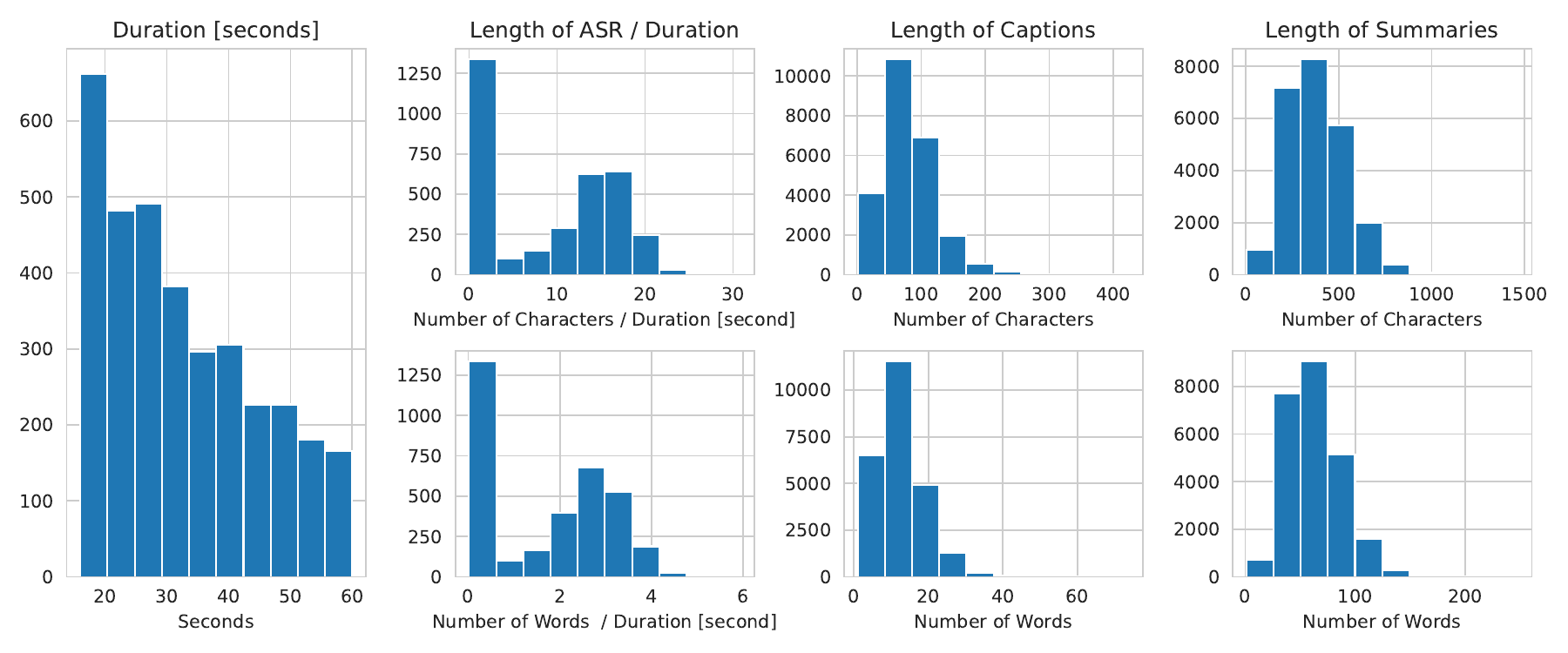} 
    \caption{{\bf Distribution of video duration, ASR content, length of captions and length of summaries of DeVAn dataset.} The \textit{Length of ASR} normalized by duration of 
    video clip indicates that our dataset covers videos ranging from no speech content to high speech content.
    Note that number of words are calculated by counting the number of white-space-seperated character groups, and may contain punctuation.}
    \label{fig:data_stats}
\end{figure}

Here we provide detailed statistics on the evaluation dataset. In particular, in \Figref{fig:data_stats} second column from left, we show that the ASR content, when normalized by duration of video, demonstrates a clear bimodal distribution, corresponding to videos with high and low ASR content selected during the two phases of annotation process. Additionally, we observe that a significant portion of video summaries are longer than 100 words, which is considerably longer than video annotations in previously available datasets.

\section{Detailed Generation Metrics for each Annotator} \label{sect:appx:human_perf_details}
Here we provide metrics comparing each annotator's caption \& summaries to all other annotators.
Each row in Table~\ref{table:human_cap_sum} correspond to metric evaluated using the given annotator's caption/summary as prediction and all others' as ground truth.
We computed both Average and Minimum values for each metric for reference, where the minimum value indicates the lower-bound of human performance.

% \usepackage{booktabs}
\begin{table}[h]
\resizebox{\linewidth}{!}{%
\begin{tabular}{@{}ccccccccc@{}}
          & \multicolumn{4}{c}{\textbf{Caption}} & \multicolumn{4}{c}{\textbf{Summary}} \\
Annotator & B4     & R       & C       & B-RT    & B4      & R       & C       & B-RT   \\ \midrule
1         & 6.5    & 33.2    & 55.3    & 51.4    & 17.3    & 35.8    & 40.0    & 56.4   \\
2         & 6.7    & 33.5    & 57.1    & 51.4    & 16.7    & 35.5    & 37.1    & 56.5   \\
3         & 7.3    & 33.0    & 57.0    & 50.9    & 16.8    & 35.3    & 41.0    & 56.4   \\
4         & 6.7    & 31.3    & 52.9    & 50.2    & 15.1    & 33.8    & 35.7    & 55.1   \\
5         & 4.5    & 29.5    & 47.1    & 48.6    & 12.4    & 32.1    & 30.9    & 53.6   \\ \midrule
Avg       & 6.3    & 32.1    & 53.9    & 50.5    & 15.7    & 34.5    & 36.9    & 55.6   \\
Min       & 4.5    & 29.5    & 47.1    & 48.6    & 12.4    & 32.1    & 30.9    & 53.6   \\ \bottomrule
\end{tabular}
}
\caption{{\bf Human Performance of Video-to-Text Generation Task.} Annotation from each annotator
is used as \textit{Prediction} and computed against \text{Ground Truth} results from all other 4 annotators. The overall metrics are then aggregated via Average and Minimum.}
\label{table:human_cap_sum}
\end{table}

To contrast the similarities between annotations of the same video across annotators to that between different annotations, we performed a text-to-text retrieval task. The motivation behind this experiment is that a good evaluation metric should produce high similarity between annotations created by different human labelers for the same video, thereby resulting in a high text-to-text retrieval performance. Guided by this intuition, we randomly selected 100 videos, and computed pairwise metrics between annotator 1's caption/summary to annotations from all other annotators. We then perform a text-to-text retrieval task using each metric and report recall@1,5,10.  As shown in Table~\ref{tab:text_to_text_ret}, for the captioning task, CIDEr and BLEURT give similar recall performance. However, for the summarization task, BLERUT metric provides much higher recall than all other N-gram based metrics, providing an indirect support for the claim that the BLEURT metric is better aligned to human preferences.

\begin{table}[h]
\resizebox{\linewidth}{!}{%
\begin{tabular}{c|ccc|ccc}
                      & \multicolumn{3}{c|}{Caption} & \multicolumn{3}{c}{Summary} \\
\multicolumn{1}{l|}{} & R@1      & R@5     & R@10    & R@1     & R@5     & R@10    \\ \hline
BLEU-4                & 45\%     & 66\%    & 71\%    & 64\%    & 80\%    & 89\%    \\
ROUGE-L               & 37\%     & 59\%    & 66\%    & 69\%    & 84\%    & 88\%    \\
CIDEr                 & 65\%     & 80\%    & 87\%    & 60\%    & 83\%    & 92\%    \\
BLEURT                & 61\%     & 87\%    & 90\%    & 83\%    & 96\%    & 99\%    \\ \hline
\end{tabular}
}
\caption{{\bf Comparison of text-to-text retrieval performances across metrics.}}
\label{tab:text_to_text_ret}
\end{table}

\begin{figure}[h]
    \centering
    \includegraphics[width=\linewidth]{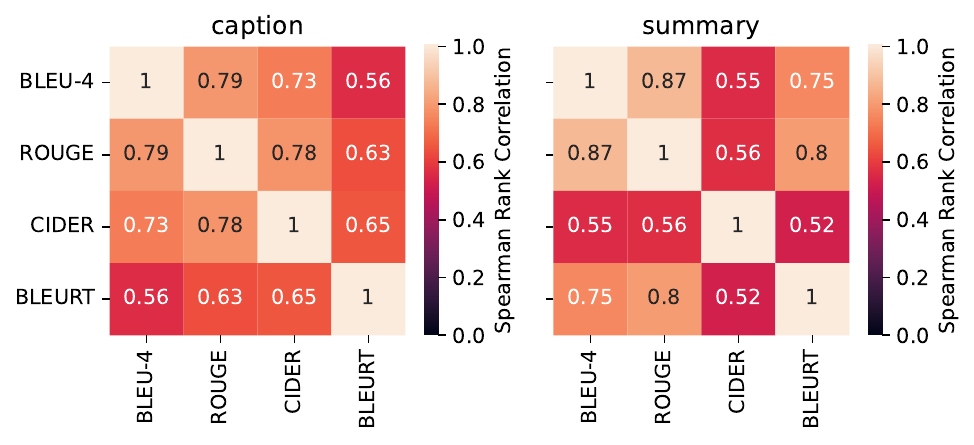} 
    \caption{{\bf Spearman Rank Correlation between evaluation metrics of human annotators.} Spearman Rank Correlation is evaluated between all human performance values for each video clip across annotators. Spearman Rank Correlaton is chosen over Pearson Correlation to emphasize the ranking consistency using different evaluation metrics.}
    \label{fig:human_metric_corr}
\end{figure}

We also computed Spearman Rank Correlation of all annotations from all annotators across evaluation metrics. As shown in \Figref{fig:human_metric_corr}. Note that Spearman Rank Correlation was chosen to emphasize rank consistency between metrics and loosens the linearity assumption implicit in Pearson Correlation.

\section{Prompt of Training Data Generation} \label{appx:sect:prompt}
The following prompt template is used for generating video summaries in training set.
Here \texttt{cc} refers to the video subtitles.

\begin{lstlisting}
summary_template = """Please write a summary in 3 to 10 sentences that 
accurately summarizes the video's content and captures 
its essence based on the following information.

Title: {{ title }}|{{ chapter_title }}

Category: {{ category }}

Description: {{ description }}

Closed Captions: {{ cc }}

SUMMARY: This video"""
\end{lstlisting}

The following prompt template is used for generation video captions in training set.
Here \texttt{cc} refers to the video subtitles.

\begin{lstlisting}
caption_template = """Please write a 1-sentence caption that 
accurately summarizes the video's content and captures 
its essence based on the following information.

Title: {{ title }}|{{ chapter_title }}

Category: {{ category }}

Description: {{ description }}

Closed Captions: {{ cc }}

ONE-SENTENCE CAPTION: This video"""
\end{lstlisting}

\section{VideoCoCa Training Details}\label{sect:appx:training_details}
VideoCoCa models are trained using 64 V100-32G GPUs with global batchsize of 256 (3 per card). Training 100K dataset for 5 epochs consumes roughly 5 hours.

Learning rate of VideoCoCa follows Linear Warmup with Cosine decay pattern, with warmup LR of 1e-7 (500 warmup steps), peak LR of 1e-5 and a minimum LR of 1e-6. 

Both VideoCoCa and VideoCoCa w/o ASR models are trained with 8 uniformly sampled input video frames for 5 epochs.

% {\color{red} hallucination not comprehensive enough}
\section{Details of Human Annotation} \label{sect:annotation}
\begin{figure}[h]
    \centering
    \includegraphics[width=\linewidth]{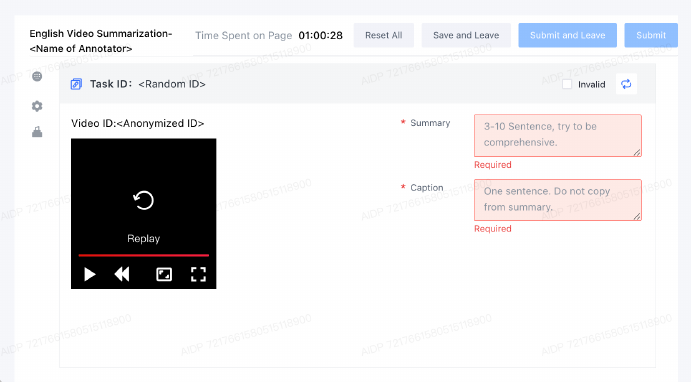}
    \caption{
    {\bf Screenshot of annotation platform.}
    }
    \label{fig:annotation_platform}
\end{figure}

Annotation of DeVAn occurred over a 10-months period, divided into multiple rounds, with each round covering 500-1500 videos. Prior to annotation, annotators were recruited and evaluated based on their performances on 200 held out videos. All accounted for, 24 human annotators (college and graduate level students) were recruited and performed their tasks on an online platform as shown in \Figref{fig:annotation_platform}. Annotators are required to fill in one caption and one summary for each video, or alternatively mark the video as invalid if it resembles a slideshow recording, is not in English, or does not have sufficient visual information to support a 3-10 sentence video description.

After each round of annotation, 20\% of videos are randomly selected for quality control independent of the original annotators. If systematic problems are detected in a batch of annotations, the entire batch is returned to annotators for revision before going through another round of quality control. In later rounds, as quality of annotation stabilized, the percentage of videos selected for independent quality control is adjusted downwards to a minimum of 7.5\%. This process is repeated until the batch at question is deemed of satisfactory quality, and every batch went through at least one round of revision.

\section{Performance of VideoCoCa model without finetuning} \label{sect:supp:videococa_no_training}
In Table~\ref{tab:videococa_no_training}, we show the performance of VideoCoCa model (CoCa-ViT-L-14) evaluated on the DeVAn summarization task. Due to the lack of instruction following capabilities of VideoCoca, it had a catastrophically low performance on the out-of-domain video summarization task without  finetuning on video summarization dataset.

\begin{table}[h]
\resizebox{\linewidth}{!}{%
\begin{tabular}{lccccc}
\toprule
\textbf{Architecture} & \textbf{Training} & \textbf{BLEU-4} & \textbf{ROUGE} & \textbf{CIDEr} & \textbf{BLEURT} \\
\midrule
VideoCoCa & DeVAn Training Set & 2.9 & 16.4 & 3.3 & 23.9 \\
VideoCoCa & No Training & 7.5 & 7.5 & 0.2 & 11.0 \\
\bottomrule
\end{tabular}
}
\caption{{\bf Video summarization performances of VideoCoCa with and without training on DeVAn training set.}}
\label{tab:videococa_no_training}
\end{table}

\section{Examples from DeVAn Dataset} \label{sect:examples}
In this section, we provide more qualitative examples from DeVAn dataset. Note that though each video has correspondingly 5 captions and summaries, only one caption and one summary is shown for brevity.
\begin{figure}[h]
    \centering
    \includegraphics[width=\linewidth]{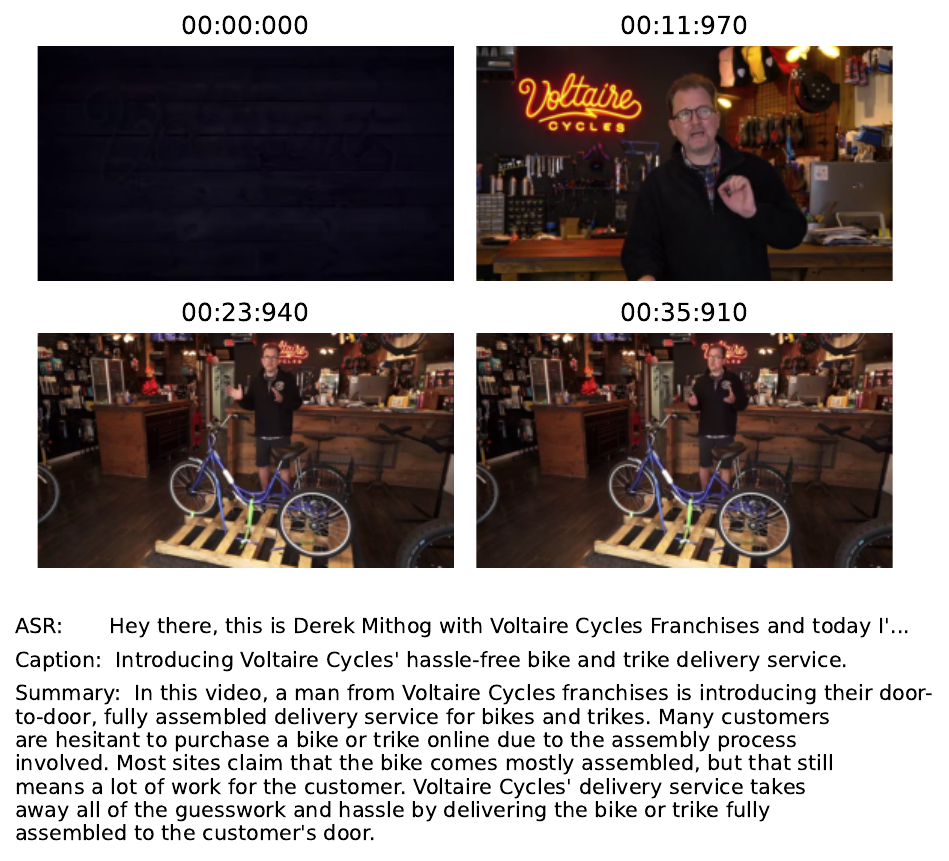}
    \caption{{\bf Example DeVAn data annotated from YouTube video zheszz4PLUw.} }
    \label{fig:example_video_zheszz4PLUw___0}
\end{figure}

\begin{figure}[h]
    \centering
    \includegraphics[width=\linewidth]{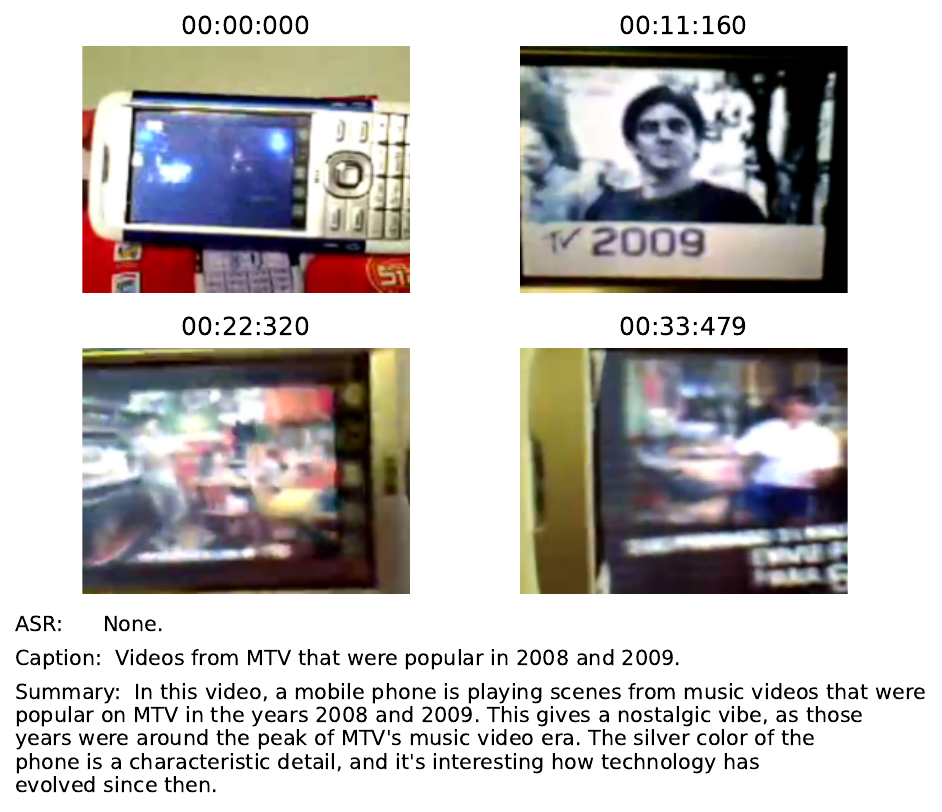}
    \caption{{\bf Example DeVAn data annotated from YouTube video -da7ZrCiupo}}
    \label{fig:example_video_-da7ZrCiupo___1}
\end{figure}

\begin{figure}[h]
    \centering
    \includegraphics[width=\linewidth]{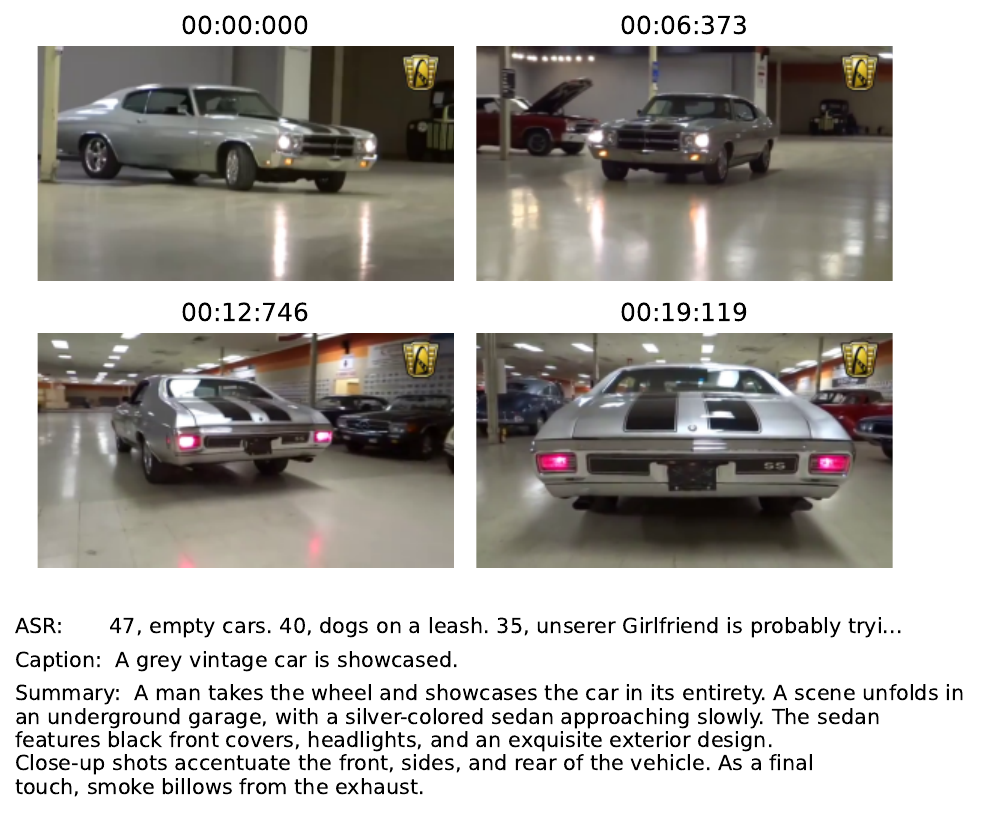}
    \caption{{\bf Example DeVAn data annotated from YouTube video 0Cmr0lSWU44.} }
    \label{fig:example_video_0Cmr0lSWU44___1}
\end{figure}

\begin{figure}[h]
    \centering
    \includegraphics[width=\linewidth]{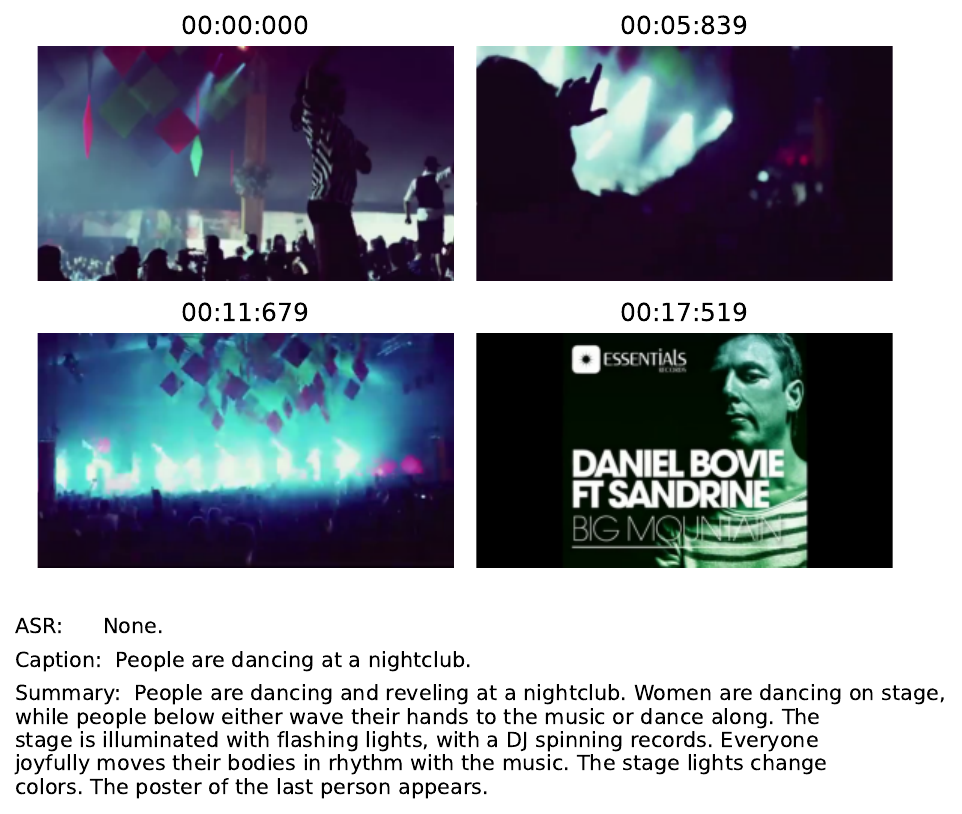}
    \caption{{\bf Example DeVAn data annotated from YouTube video 6CLbKXgWIvU.} }
    \label{fig:example_video_6CLbKXgWIvU___1}
\end{figure}

\begin{figure}[h]
    \centering
    \includegraphics[width=\linewidth]{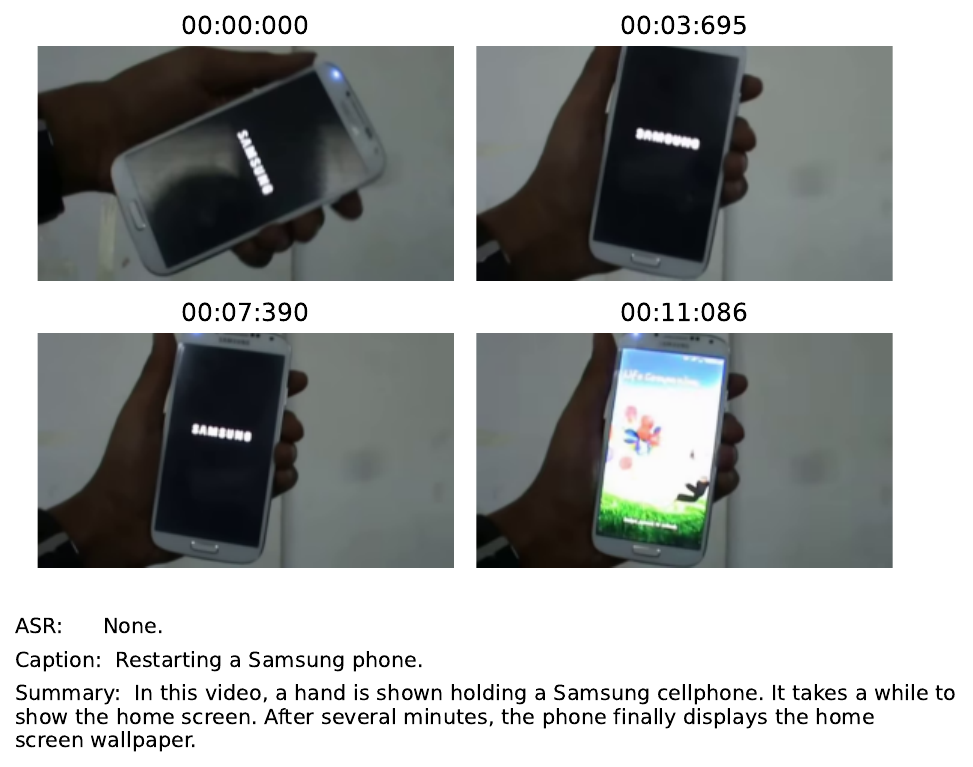}
    \caption{{\bf Example DeVAn data annotated from YouTube video 38jPK8lRlb0.} }
    \label{fig:example_video_38jPK8lRlb0___3}
\end{figure}

\begin{figure}[h]
    \centering
    \includegraphics[width=\linewidth]{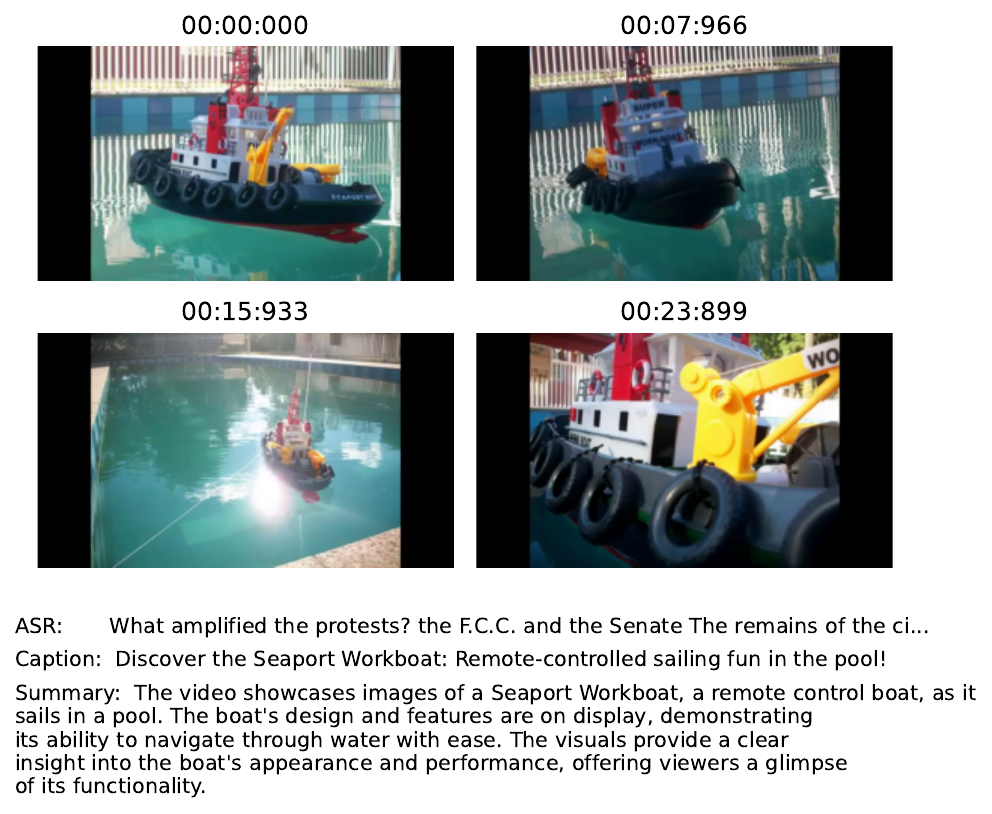}
    \caption{{\bf Example DeVAn data annotated from YouTube video cb\_HvxX80sE.}}
    \label{fig:example_video_cb_HvxX80sE___2}
\end{figure}

\begin{figure}[h]
    \centering
    \includegraphics[width=\linewidth]{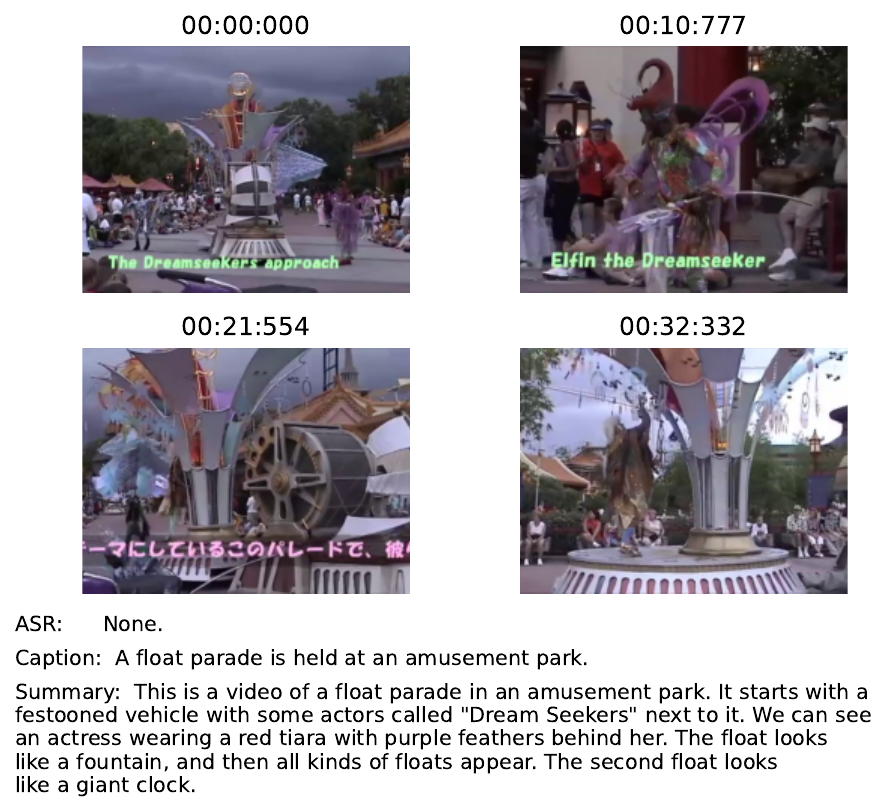}
    \caption{{\bf Example DeVAn data annotated from YouTube video HVmx56OSjDE.} }
    \label{fig:example_video_HVmx56OSjDE___1}
\end{figure}

\begin{figure}[h]
    \centering
    \includegraphics[width=\linewidth]{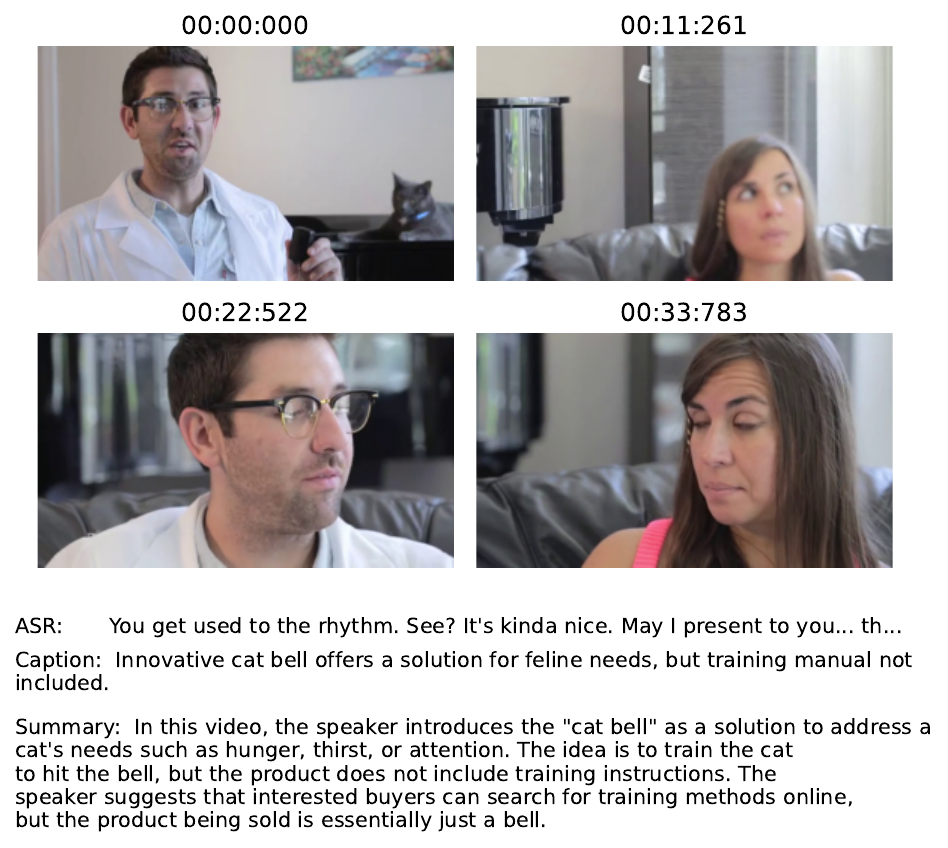}
    \caption{{\bf Example DeVAn data annotated from YouTube video Hvo6APy2f5A.} }
    \label{fig:example_video_Hvo6APy2f5A___2}
\end{figure}

\begin{figure}[h]
    \centering
    \includegraphics[width=\linewidth]{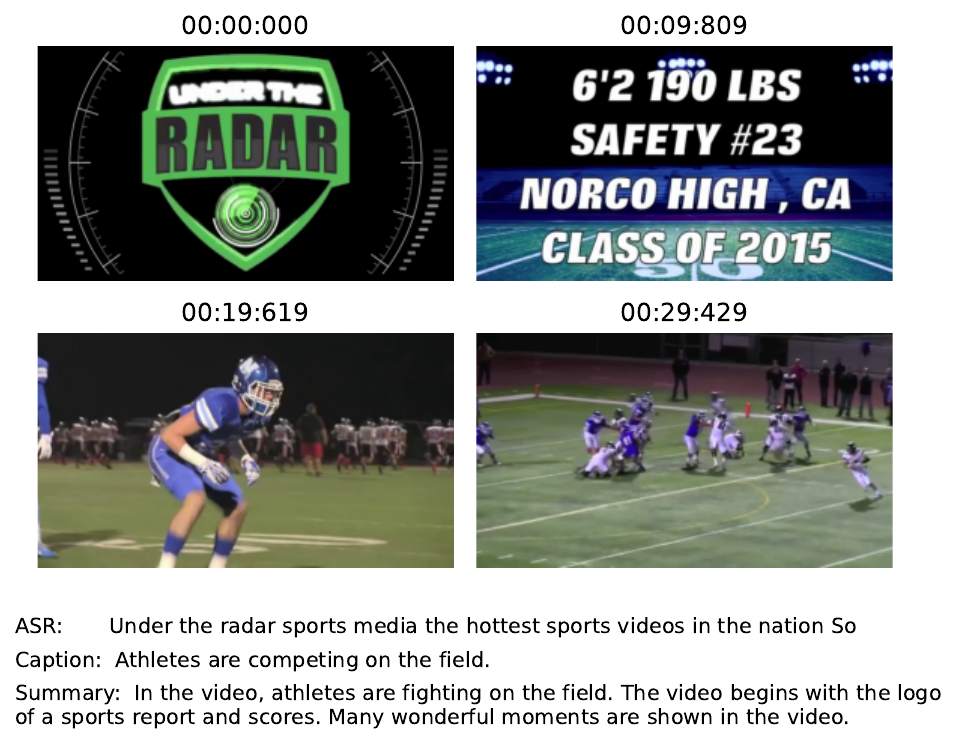}
    \caption{{\bf Example DeVAn data annotated from YouTube video LCmfVX2S8zs.} }
    \label{fig:example_video_LCmfVX2S8zs___1}
\end{figure}

\begin{figure}[h]
    \centering
    \includegraphics[width=\linewidth]{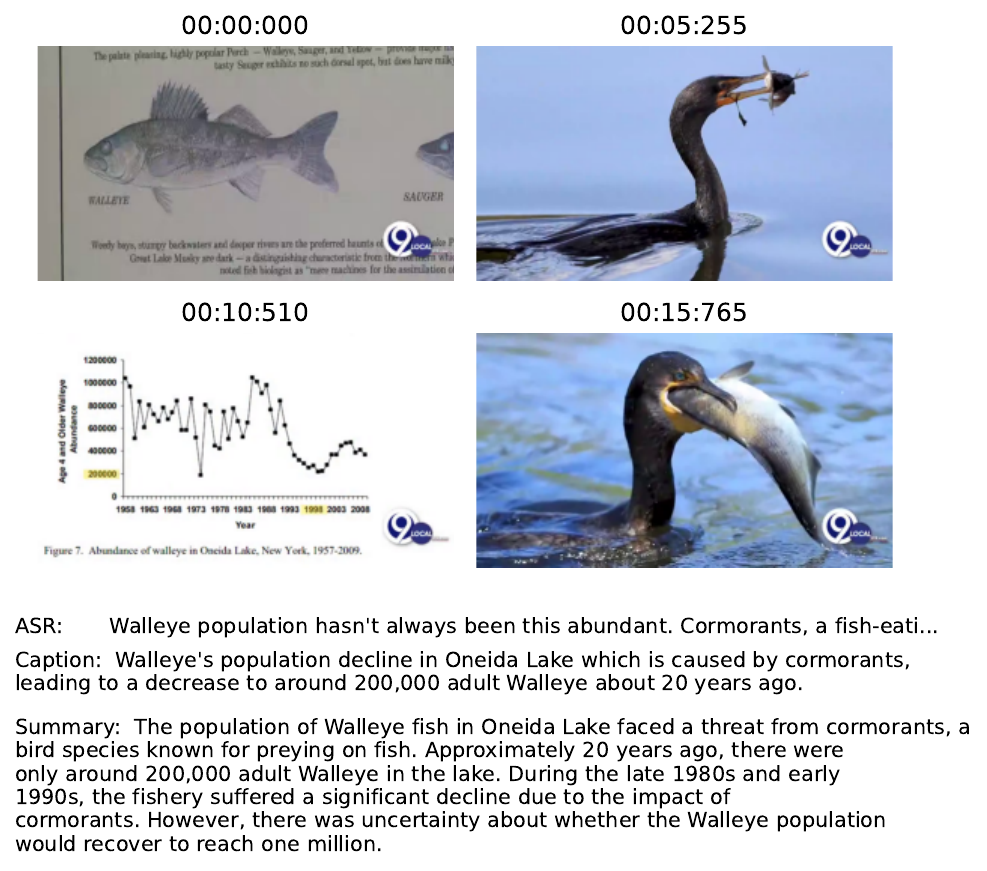}
    \caption{{\bf Example DeVAn data annotated from YouTube video MhZRLhwcsJg.}}
    \label{fig:example_video_MhZRLhwcsJg___1}
\end{figure}

\begin{figure}[h]
    \centering
    \includegraphics[width=\linewidth]{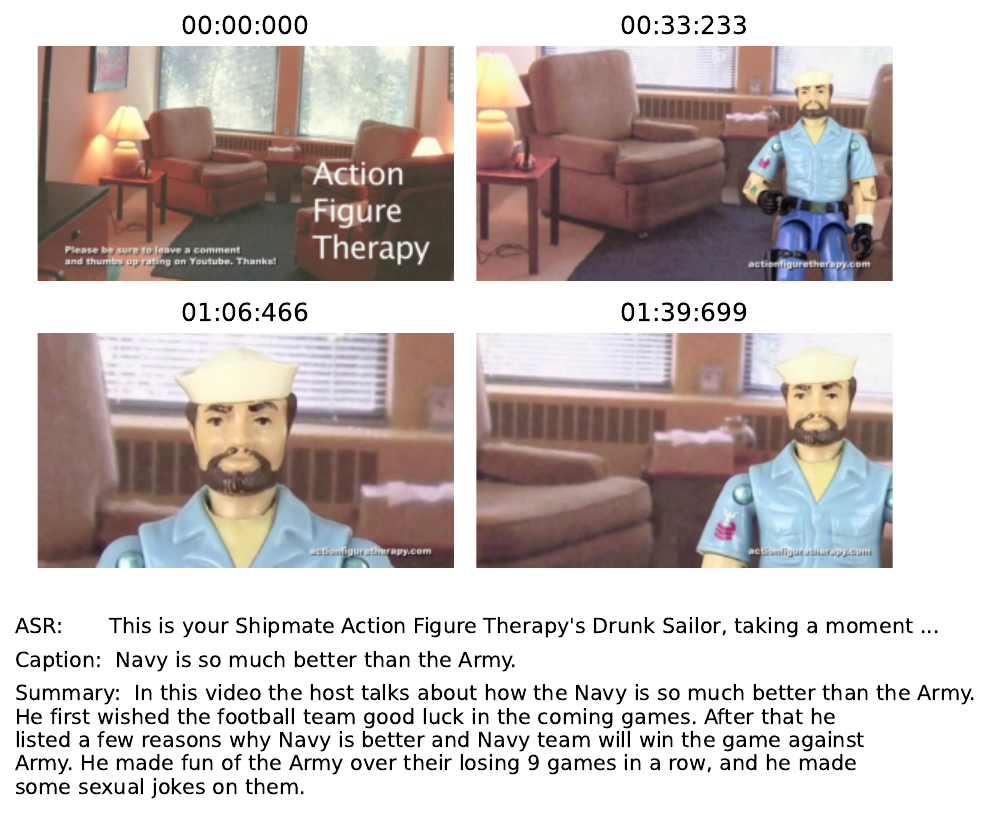}
    \caption{{\bf Example DeVAn data annotated from YouTube video pKC23tWTi-k.}}
    \label{fig:example_video_pKC23tWTi-k}
\end{figure}

\section{Pathological Examples Captions Generated from Video-Language Models} \label{sect:pathological_captions}
We noted that models like Video-LLaMA2-Instruct had very low CIDEr scores but high BLEURT scores. 
In this section, we show some pathological examples where model generated responses show close to 0 CIDEr score but very high BLEURT scores.

\begin{figure}[h]
    \centering
    \includegraphics[width=\linewidth]{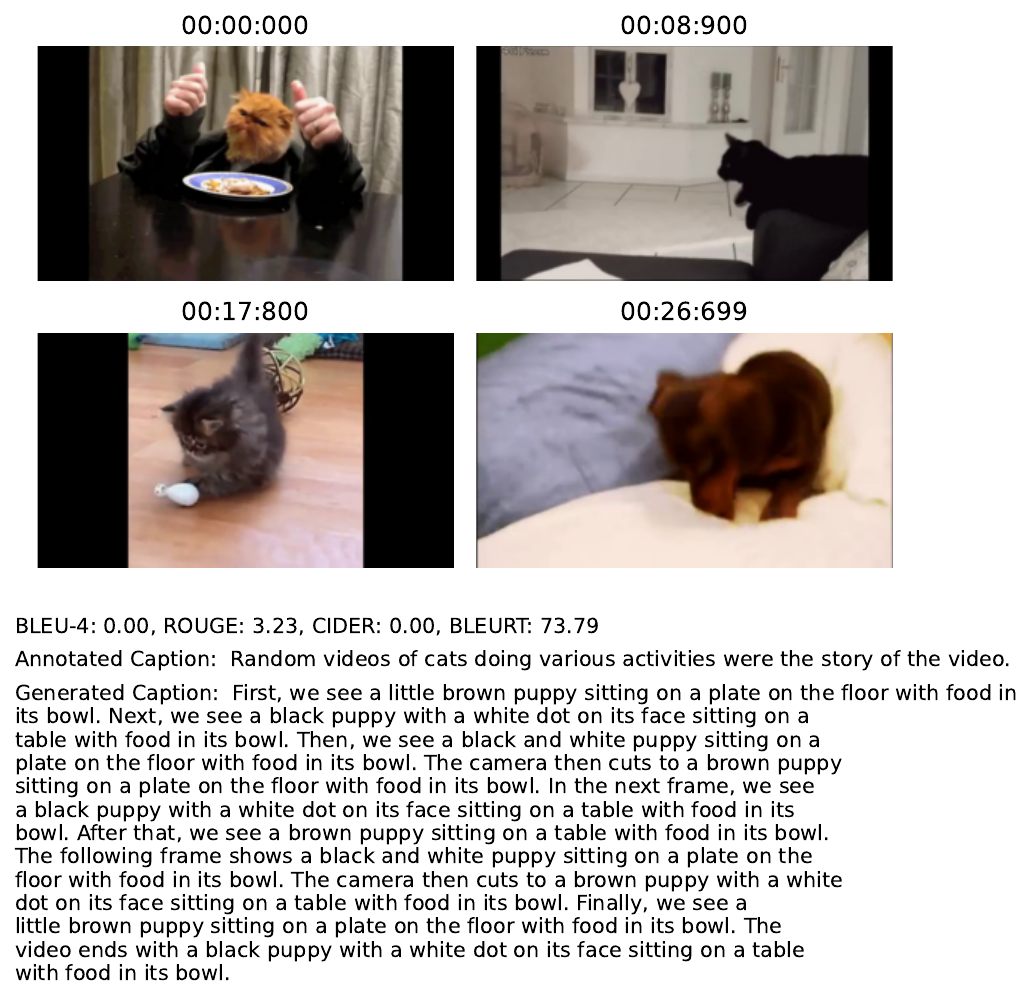}
    \caption{{\bf Pathological example caption data generated by Video-LLaMA2-Instruct for video clip take from YouTube video 1iJwajSfE6g.}}
    \label{fig:pathological_1iJwajSfE6g___1}
\end{figure}

\begin{figure}[h]
    \centering
    \includegraphics[width=\linewidth]{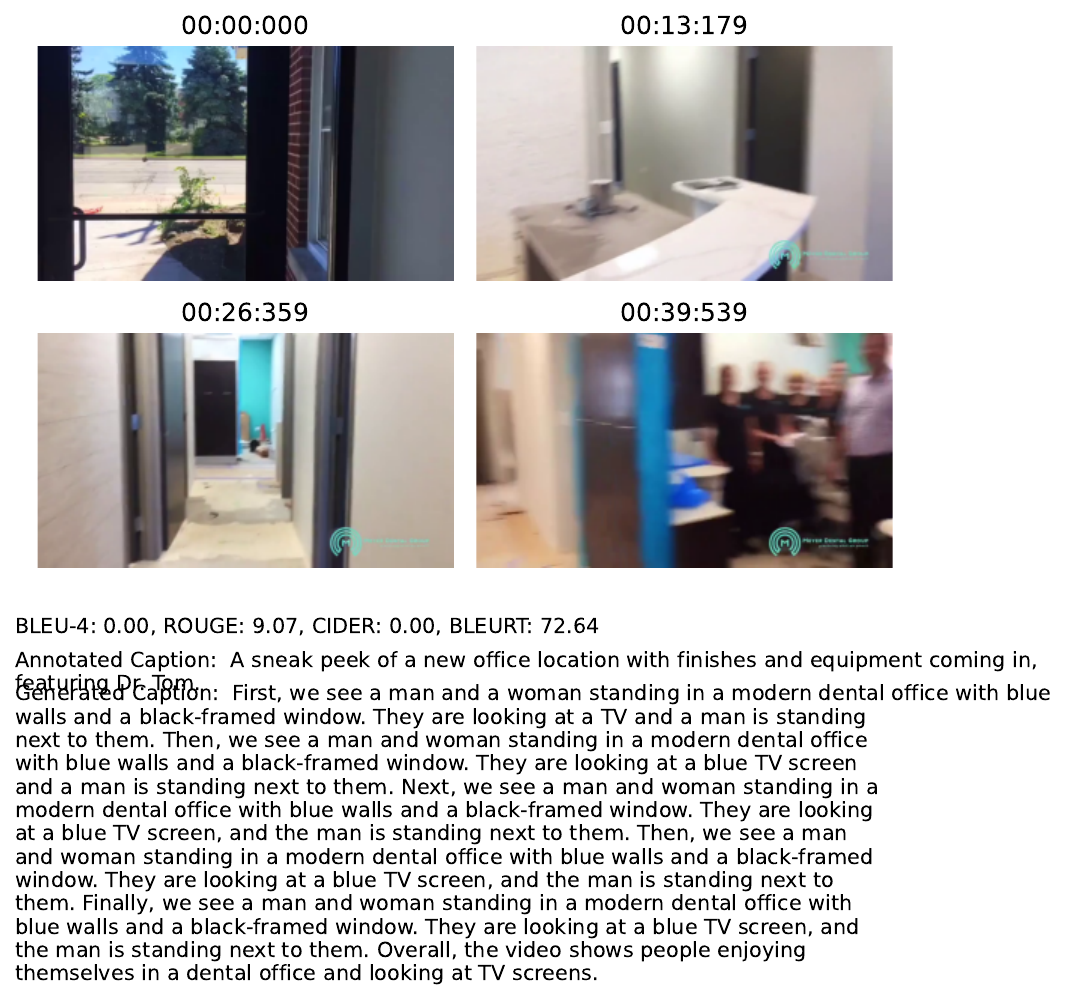}
    \caption{{\bf Pathological example caption data generated by Video-LLaMA2-Instruct for video clip take from YouTube video 1sRqzVs3etI.}}
    \label{fig:pathological_1sRqzVs3etI}
\end{figure}

\begin{figure}[h]
    \centering
    \includegraphics[width=\linewidth]{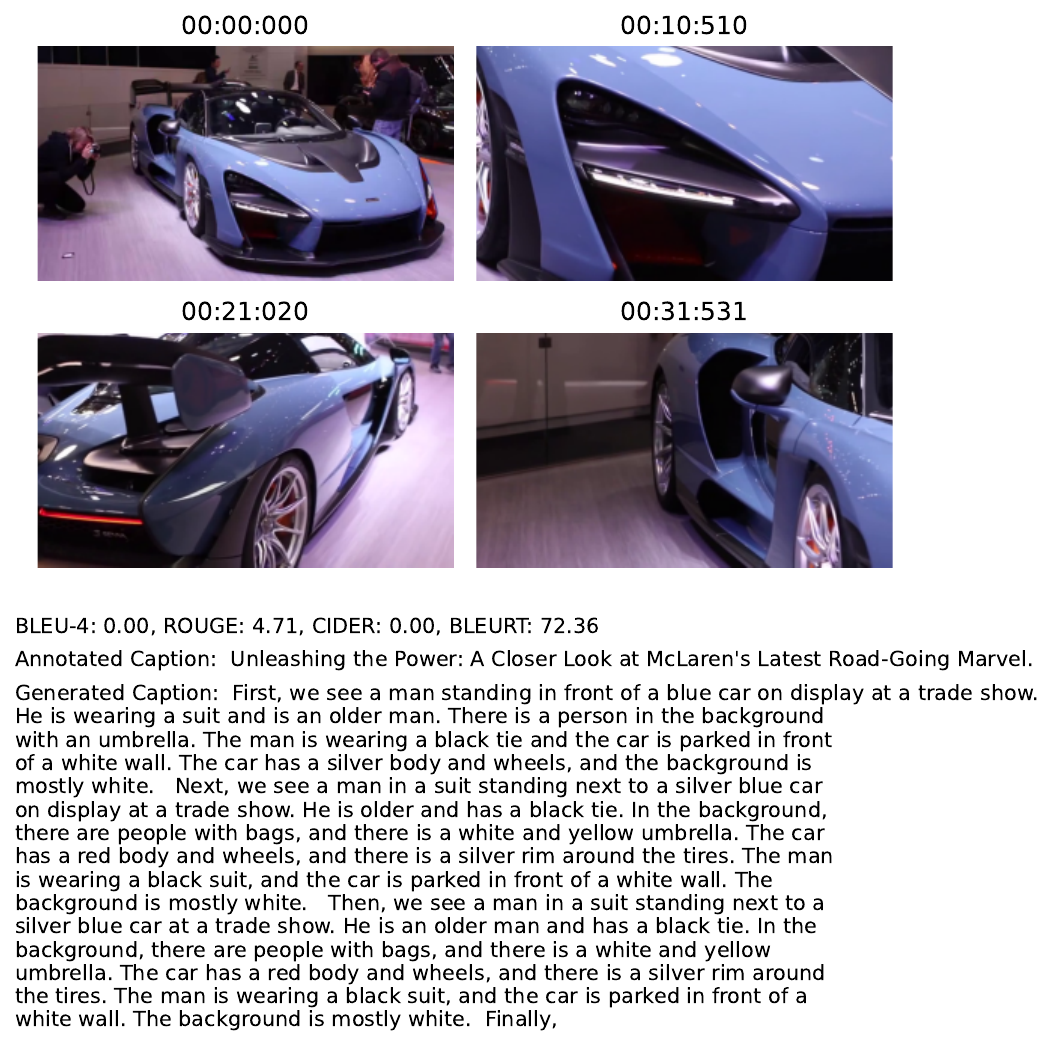}
    \caption{{\bf Pathological example caption data generated by Video-LLaMA2-Instruct for video clip take from YouTube video \_6st0uhrcBZU.}}
    \label{fig:pathological_6st0uhrcBZU}
\end{figure}

\begin{figure}[h]
    \centering
    \includegraphics[width=\linewidth]{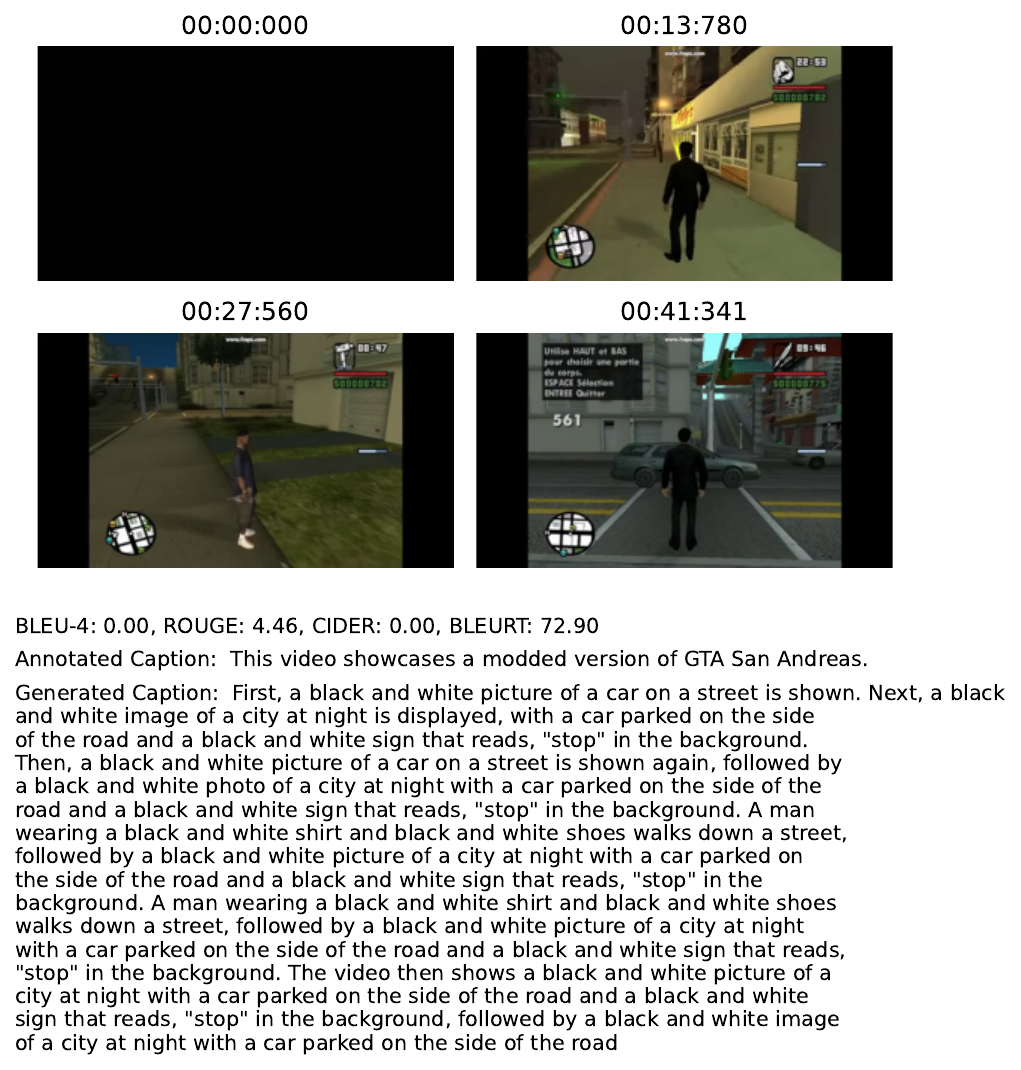}
    \caption{{\bf Pathological example caption data generated by Video-LLaMA2-Instruct for video clip take from YouTube video eoIJuYUA0UI.}}
    \label{fig:pathological_eoIJuYUA0UI}
\end{figure}